\documentclass{article} %
\usepackage{preprint,times}

\usepackage{hyperref}
\hypersetup{
  pdftitle={The Intrusion Pathway: How Non-Evidential Context Degrades Language Model Performance},
  pdfauthor={Dikshant Kukreja, Gautam Gupta, Avni Mittal, Avinash Anand, Kritarth Prasad, Zhengkui Wang, Timothy Liu, Rajiv Ratn Shah, Aik Beng Ng, Simon See, Erik Cambria}
}
\usepackage{url}

\usepackage{latexsym}
\usepackage[T1]{fontenc}
\usepackage[utf8]{inputenc}
\usepackage{microtype}
\usepackage{inconsolata}
\usepackage{graphicx}
\usepackage{booktabs}
\usepackage{array}
\usepackage{multirow}
\usepackage{amsmath}
\usepackage{amssymb}
\usepackage{placeins}
\usepackage[most]{tcolorbox}
\usepackage{tikz}
\usepackage{wrapfig}
\makeatletter
\newcommand{\wrapfill}{\par
  \ifx\parshape\WF@fudgeparshape
    \nobreak
    \vskip-\baselineskip
    \vskip\c@WF@wrappedlines\baselineskip
    \allowbreak
    \WFclear
  \fi}
\makeatother

\renewcommand{\floatpagefraction}{0.7}

  \definecolor{cGold}{HTML}{009E73}  %
  \definecolor{cCF}{HTML}{D62728}    %
  \definecolor{cRel}{HTML}{1F77B4}   %
  \definecolor{cIrrS}{HTML}{E6B400}  %
  \definecolor{cRand}{HTML}{984EA3}  %
  \definecolor{cIrrM}{HTML}{56B4E9}
  \newcommand{\tbarhalfval}{2.4}   %
  \newcommand{\tbarspan}{5.0}      %
  \newcommand{\tbar}[2]{%
    \begin{tikzpicture}[baseline=-0.55ex, x=1cm, y=1cm]
      \def\tbh{0.16}
      \pgfmathsetmacro{\tbx}{\tbarhalfval*(#2)/\tbarspan}%
      \pgfmathsetmacro{\tbr}{min(\tbh,abs(\tbx))}%
      \useasboundingbox (-\tbarhalfval,-0.21) rectangle (\tbarhalfval,0.21);
      \ifdim \tbx pt>0pt
        \fill[#1] (0,-\tbh) -- ({\tbx-\tbr},-\tbh)
          arc[start angle=-90, end angle=90, x radius=\tbr, y radius=\tbh]
          -- (0,\tbh) -- cycle;
        \node[anchor=west, font=\bfseries, inner sep=2.5pt,
              text=black!75]
          at (\tbx,0) {\pgfmathprintnumber[print sign,fixed,precision=1]{#2}};
      \else
        \fill[#1] (0,-\tbh) -- ({\tbx+\tbr},-\tbh)
          arc[start angle=-90, end angle=-270, x radius=\tbr, y radius=\tbh]
          -- (0,\tbh) -- cycle;
        \node[anchor=east, font=\bfseries, inner sep=2.5pt,
              text=black!75]
          at (\tbx,0) {\pgfmathprintnumber[print sign,fixed,precision=1]{#2}};
      \fi
      \draw[black!60, line width=0.5pt] (0,-0.24) -- (0,0.24);
    \end{tikzpicture}%
  }

\title{\raggedright The Intrusion Pathway: How Non-Evidential Context Degrades Language Model \mbox{Performance}}

\newcommand{\websiteurl}{https://dk4248.github.io/the-intrusion-pathway/}
\newcommand{\codeurl}{https://github.com/dk4248/the-intrusion-pathway}

\author{%
  {\small Dikshant Kukreja$^{1,*}$, Gautam Gupta$^{2,*}$, Avni Mittal$^{3}$, Avinash Anand$^{4}$, Kritarth Prasad$^{4}$,} \\[3pt]
  {\small\bfseries Zhengkui Wang$^{4}$, Timothy Liu$^{5}$, Rajiv Ratn Shah$^{6}$, Aik Beng Ng$^{5}$, Simon See$^{5}$, Erik Cambria$^{7}$} \\[4pt]
  {\footnotesize $^{1}$TurboML \quad $^{2}$Indraprastha Institute of Information Technology, Delhi \quad $^{3}$Independent Researcher} \\[1pt]
  {\footnotesize $^{4}$Singapore Institute of Technology \quad $^{5}$NVIDIA \quad $^{6}$Indian Institute of Technology, Kanpur} \\[1pt]
  {\footnotesize $^{7}$Nanyang Technological University \quad $^{*}$Equal contribution} \\[4pt]
  {\footnotesize \paperlink{\websiteurl}{\websiteicon}{Website}} \\[1pt]
  {\footnotesize \paperlink{\codeurl}{\codeicon}{Code}}
}

\begin{document}
\maketitle
\enlargethispage{24bp}
\begin{abstract}
Language models lose accuracy when a prompt contains non-evidential context: irrelevant statements from unrelated domains, or random off-distribution tokens, with no competing answer and no semantic overlap with the query. Such noise pervades retrieval-augmented, tool-using, and long-context deployments, yet the failure has lacked a mechanistic account. We trace the corruption to the \emph{Intrusion Pathway}, which starts with MLPs enriching both query-subject and irrelevant-token representations, parroting heads carry prefix-token content to the prediction position, and relational readers largely avoid unrelated sources. Nevertheless, preserved subject enrichment and selective reading do not prevent intrusion in the broader computation. We then localize part of this interference to a sparse set of attention heads, which we call \emph{intrusion heads}. Their prefix-derived writes promote distractor-related content, while their reduced support for the correct answer arises primarily from weaker non-prefix contributions rather than direct suppression by the prefix. The same heads transfer across tasks and distractor types and are largely weight-disjoint from the heads that integrate relevant contextual evidence, enabling a class-conditional intervention. Replacing the corrupted last-token residual with its clean counterpart, an oracle probe that requires a matched clean forward pass, recovers 97.6\% on average of the distractor-induced drop across four downstream tasks. A deployable mask level knockout of about ten heads, roughly 1\% of the model’s attention heads, recovers 57\% at bounded cost to helpful context.

\end{abstract}

\section{Introduction}

A current-day frontier language model or agent rarely sees a bare question. A user's prompt arrives bundled with retrieved passages from a vector store \citep{lewis2020rag}, tool-call results from a browsing or search API \citep{schick2023toolformer}, conversation snippets, and traces from agentic sub-loops \citep{yao2023react}. The model is expected to weld this heterogeneous bundle into a single coherent answer, and the bundle is rarely clean: context closes real factual gaps that parametric memory cannot reach on its own \citep{mallen2023trust}, but the same channel just as routinely carries pieces that share none of the answer's content. The pieces of this mixture fall into a now-standard five-way partition---\emph{golden}, \emph{counterfactual}, \emph{related}, \emph{irrelevant}, and \emph{random} context---established across the retrieval-augmentation, knowledge-conflict, and contextual-distraction literatures \citep{cuconasu2024noise,niu2025llama,fang2024noise,wu2025pandora}.

For a query $q$ with answer $a^\star$, the usage above partitions any context $c$ into five disjoint classes, each instantiated in Table~\ref{tab:taxonomy} on the running query \emph{``The capital of Germany is \_\_\_''} ($a^\star = $ Berlin). \emph{Golden} context contains $a^\star$ and closes factual gaps the model would otherwise miss \citep{mallen2023trust}; its extraction machinery is the best-mapped circuit in the field \citep{wang2023ioi,mcdougall2023copy}. \emph{Counterfactual} context asserts a coherent competing answer via the query's own relation; it is the substrate of the knowledge-conflict literature \citep{longpre2021entity} and has been localized at every grain, from mid-layer FFNs to individual memory-vs-context heads. \emph{Related} context shares the relational schema of $q$ but asserts about a different entity, functioning as a one-shot demonstration of the task: this is the canonical in-context learning regime of induction heads and task/function vectors \citep{olsson2022induction,hendel2023task}. \emph{Irrelevant} context is a coherent statement from an unrelated domain---no shared entity, relation, or topic with $q$---and contains one or more salient non-gold (distractor) tokens. Such context can substantially reduce answer accuracy \citep{shi2023distracted,wu2024easily}. \emph{Random} context is an arbitrary token or short string with no semantic content, a control for whether the model keys on distribution rather than content~\mbox{\citep{cuconasu2024noise}}.

Among these five classes, irrelevant context is the least understood mechanistically. \citet{shi2023distracted} document the accuracy loss it causes, and \citet{niu2025llama} identify heads responsible for contextual entrainment, the promotion of tokens that appear in the context. Consider again ``The capital of Germany is,'' whose correct continuation is ``Berlin.'' Prepending ``Computers are useful'' neither supports this answer nor proposes another capital, yet it lowers the gold token's logit. Promoting a distractor can reduce the gold token's probability through competition, but does not by itself establish why the gold token's own logit decreases. We therefore ask two questions: how does content that supplies no evidence about the answer reduce support for the gold token, and can the components that carry this interference be separated from those that let the model use golden context? Answering these questions would clarify why irrelevant context interferes with transformer computation and how much robustness can be recovered through a sparse, training-free intervention, rather than relying on noise-aware training or additional context-handling procedures \citep{fang2024noise,yoran2024making,asai2024selfrag,yu2024chainofnote}.\looseness=-1

We answer the first question by tracing how context is represented and propagated, then localizing the writes associated with lost answer support. In \S\ref{sec:pathway}, our analysis of the \textbf{Intrusion Pathway} follows MLP enrichment, parroting, and relational reading. The measured enrichment of the query subject is preserved under irrelevant context, while irrelevant-token representations are also enriched. Parroting heads carry prefix-token content to the last position, and relational readers largely avoid unrelated sources. Nevertheless, restoring the broader middle block’s writes recovers answer support while also removing benefits from golden context, identifying an effective but non-selective intervention site.

In \S\ref{sec:pathway-intrusion}, we identify \textbf{intrusion heads}, whose attention to irrelevant context coincides with reduced direct support for the correct answer. Their prefix-derived writes promote distractor-related content, whereas their reduced gold support comes primarily from weaker contributions from the query. This localizes a component of the interference to changes in the heads’ source contributions, rather than treating copying alone as the mechanism of answer degradation.

For the second question, intrusion heads are largely distinct from golden-context retrieval heads (\citealp{wu2024retrieval}; \S\ref{sec:pathway-intrusion}). Masking prefix attention in a fixed, retrieval-disjoint set of about ten heads per model, roughly $1\%$, recovers $57\%$ of distractor-induced accuracy loss on average across four downstream tasks while largely preserving golden-context benefits. The set is selected once on factual recall, and the mask requires no matched clean forward pass at inference. Oracle replacement of the last-position residual with its matched clean counterpart recovers $97.6\%$ on average (\S\ref{sec:downstream}).

\begin{table}[!t]
\centering
\small
\setlength{\tabcolsep}{4pt}
\renewcommand{\arraystretch}{1.10}
  \begin{tabular}{@{}>{\raggedright\arraybackslash}p{2.6cm}
                    >{\raggedright\arraybackslash}p{4.8cm}
                    >{\raggedright\arraybackslash}p{5.8cm}@{}}
                    \toprule
  \textbf{Regime} & \textbf{Context} & \hspace{\tbarhalfval cm}\makebox[0pt][c]{\textbf{Gold-logit shift}} \\
  \midrule
Golden &
  ``Germany's capital city is Berlin.'' &
  \tbar{cGold}{4.8} \\
  Counterfactual &
  ``The capital of Germany is Munich.'' &
  \tbar{cCF}{-2.6} \\
  Related &
  ``The capital of France is Paris.'' &
  \tbar{cRel}{1.8} \\
  Irr.\ (single token) &
  ``Computers are useful.'' &
  \tbar{cIrrS}{-1.1} \\
  Irr.\ (multi-token) &
  ``Computers perform ... efficiently.'' &
  \tbar{cRand}{-2.3} \\
  Random &
  ``Calculator.'' &
  \tbar{cIrrM}{-0.2} \\

\bottomrule
  \end{tabular}%
\caption{Context regimes for the running query ``The capital of Germany is \_\_\_'' (\(a^\star=\text{Berlin}\)). The \textbf{Gold-logit shift} column shows the gold token (Berlin) logit change caused in nats by that regime. ``Irr.'' abbreviates Irrelevant; the full multi-token prefix is ``Computers perform mathematical simulations efficiently.''}
\label{tab:taxonomy}
\end{table}

\section{Experimental setting and background}
\label{sec:data}

For next-token circuit analysis, we compare a query \(q\) alone (\(\mathrm{nctx}\)) with the prefixed input \(c\parallel q\), holding the query, scored gold token \(g\), and model fixed.
\textbf{Task data.} The analyses for intrusion is characterized on two circuit benchmarks---factual recall and binding/retrieval. Factual recall uses LRE source--target pairs rendered through relation-specific templates \citep{hernandez2023lre}. Binding/retrieval uses indirect object identification (IOI) templates \citep{wang2023ioi}. Irrelevant prefixes are approximately 12-token sentences (\texttt{short\_irr}) or approximately 100-token WikiText-103 paragraphs (\texttt{long\_irr}; \citealp{wikitext103}). Long paragraphs are filtered for subject, answer, and related lexical or semantic overlap. Appendix~\ref{app:data} provides construction and filtering details.
\textbf{Models.} We use five open-weight base models---Llama-3.1-8B \citep{grattafiori2024llama3}, Mistral-7B-v0.1 \citep{jiang2023mistral}, Qwen2.5-7B \citep{yang2024qwen25}, Pythia-6.9B \citep{biderman2023pythia}, and GPT-2 XL \citep{radford2019language}---through TransformerLens \citep{nanda2023tlens}. Downstream evaluation is described separately.
\textbf{Metrics.} Let $\ell_t(q)$ and $P(t\mid q)$ denote token $t$'s logit and next-token probability at the last prompt position. We measure the gold-logit shift $\Delta\ell_g(c;q)=\ell_g(c\mathbin{\|}q)-\ell_g(q)$ and token log-probability shift $\Delta_t(c;q)=\log P(t\mid c\mathbin{\|}q)-\log P(t\mid q)$. Negative $\Delta\ell_g$ indicates that the prefix lowers the gold token's logit; negative $\Delta_g$ indicates that it lowers the probability assigned to that token. For a designated distractor token $d$ appearing in the prefix, positive $\Delta_d$ indicates distractor entrainment. A decrease in gold probability can also result from increased competing logits, even when the gold logit is unchanged. We report paired changes averaged over the full dataset for each task, model, and evaluated prefix condition. Downstream, we report accuracy and recovery: the percentage of distractor-induced accuracy loss restored by an intervention.\looseness=-1

\paragraph{Notation.} END is the last prompt position, and $x_p^{(L)}$ is the residual-stream vector at position $p$, layer $L$. Let $u_t=W_U[:,t]$ be token $t$'s unembedding direction. Head $h$'s QK circuit sets attention weights $\alpha_p=A_h[\mathrm{END},p]$, while its OV circuit writes
\begin{equation}
  o_h=\sum_p\alpha_p W_O^{(h)}W_V^{(h)}x_p^{(L)},
\label{eq:setup-ov}
\end{equation}
with normalization and bias terms suppressed. We use the raw projection $\mathrm{DLA}_h(t)=\langle o_h,u_t\rangle$ for direct logit attribution, measuring directional alignment rather than an intervention-based causal effect.

We introduce these copying behaviours here to specify what it means for a head to write prefix-token content into the END residual, before examining the corresponding heads and their effects.

\paragraph{Induction versus parroting.} Induction is prefix-matched copying: in the canonical pattern $[A][B]\ldots[A]\rightarrow[B]$, the repeated $[A]$ supplies a cue to attend to the earlier continuation $[B]$ and promote it as the next token \citep{olsson2022induction}. Parroting instead promotes previously encountered tokens without requiring that matching-prefix cue. These terms describe behaviours, not mutually exclusive head populations: \citet{goldowsky2023localizing} describe a head that contributes to both. We use parroting for the antecedent-independent copying contribution studied here.

\enlargethispage{24bp}
\begin{wrapfigure}{r}{0.46\textwidth}
\vspace{-\intextsep}
\centering
\includegraphics[width=\linewidth]{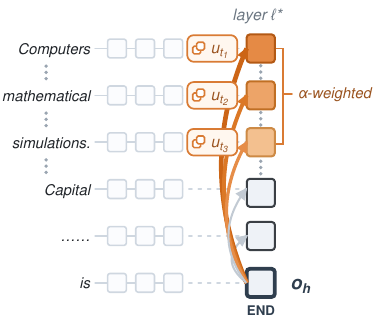}
\caption{Diffuse parroting at layer $\ell^*$: attention from END spreads over several prefix tokens, whose unembedding directions are written into the END residual $o_h$.}
\label{fig:diffuse-parroting}
\vspace{-\intextsep}
\end{wrapfigure}
\paragraph{Single-source and diffuse parroting.}
Copying may concentrate on one token or span several, such as
\mbox{``Computers''}, ``mathematical'', and ``simulations'' in an
irrelevant sentence. We call the latter \emph{diffuse parroting},
descriptively rather than as a new mechanism (Figure~\ref{fig:diffuse-parroting}).
For prefix-content positions $\mathcal{P}_{\mathrm{content}}$,
excluding BOS, stopwords, and query repetitions, define the
attention-weighted identity reference and decompose the
corresponding write as
\begin{equation}
\begin{aligned}
u_*^{\mathrm{attn}}
    &= \sum_{p\in\mathcal{P}_{\mathrm{content}}}
       \alpha_p u_{t_p}, \\
o_h^{\mathrm{content}}
    &= c_h u_*^{\mathrm{attn}} + r_h .
\end{aligned}
\label{eq:diffuse-parroting-reference}
\end{equation}
Here, $c_h>0$ is a shared copy gain, and $r_h$ is the
deviation of the actual content write from the identity-only
reference; it is not assumed small or orthogonal.
When one source dominates the mixture, the reference reduces
approximately to $c_h\alpha_{p_s}u_{t_{p_s}}$.
Multiple sources need not receive equal attention.
Appendix~\ref{app:parroting-reference} gives the derivation
and assumptions; Appendix~\ref{app:diffuse-reconstruction}
reports the held-out reconstruction test.
\wrapfill %

\section{Tracing the Intrusion Pathway}
\label{sec:pathway}
\label{sec:pathway-parroting} %

We seek to explain how irrelevant context changes the computation of an answer already determined by the query. Recall ``The capital of Germany is \_\_\_'' and the two prefixes in our running example: ``Computers are useful'' and ``Germany's capital city is Berlin.'' The former supplies no evidence about the answer, while the latter states it explicitly. Their different effects raise a question about how the model handles contextual information---not just which words it reads, but what it represents about them, carries to the answer position, and uses in the final prediction.

\textbf{Subject and prefix enrichment.} A natural place to look is the subject itself: does the sentence about computers disrupt information associated with Germany? Subject-position MLP writes make related attributes available in the subject representation \citep{geva2023dissecting}. Their support for independently defined subject attributes, relative to another subject's attributes, provides a measure of this enrichment. These writes favor their own subject's attributes, and the paired preservation test shows that this measured signal remains effectively unchanged after an irrelevant prefix is added (Appendix~\ref{sec:app-subject}). Lower support for ``Berlin'' therefore need not reflect the disappearance of information associated with Germany.

Enrichment is not confined to the queried subject. MLP writes at irrelevant-token positions also favor words associated with those tokens over another distractor's associated words (Appendix~\ref{sec:app-distractor}). In our running example, ``Germany'' and ``Computers'' thus differ in their relevance to the question, not in whether their representations can contain associated information.

The prefix effect also appears in the IOI-like setting. In ``Springs compress under load. When Mary and John went to the store, John gave a drink to,'' answering the query requires identifying the participants' roles rather than recalling their factual attributes. The diagnostic detects no comparable factual-attribute enrichment at the tested name positions, but associated information still appears at the irrelevant-prefix positions. Prefix enrichment therefore also occurs when the query does not ask for a factual association.

\enlargethispage{24bp}
\begin{wrapfigure}{r}{0.64\textwidth}
\vspace{-\intextsep}
\centering
\includegraphics[trim=10.0 73.4 12.6 75.2, clip, angle=-90, width=\linewidth]{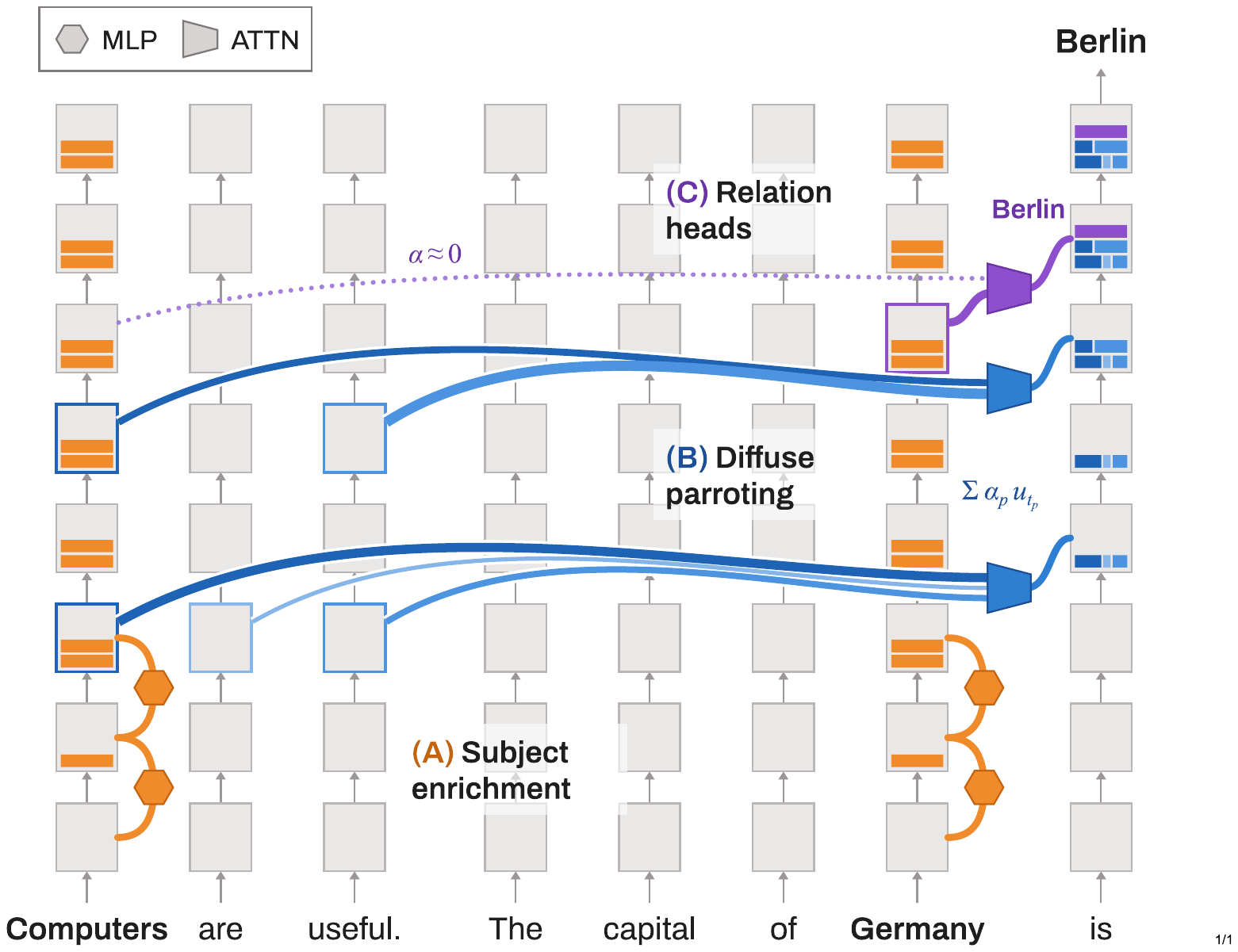}
\setlength{\abovecaptionskip}{2pt}
\caption{Irrelevant content can coexist with answer-supporting information at the prediction position: avoiding unrelated sources does not, by itself, remove their earlier contributions to the residual stream.}
\label{fig:pathway}
\end{wrapfigure}
Deleting the complete MLP writes does not isolate these associations. Cancelling subject-position writes lowers the final gold logit under both bare and irrelevant inputs, and the early-layer reduction also persists under activation replacement (Appendix~\ref{sec:app-mlp-intervention}). These interventions establish the importance of the complete writes, rather than attributing their entire effect to the association signal measured within them.

\textbf{Parroting to END.} How does information from a source such as ``Computers'' reach END, the position from which the answer is predicted? For copying to occur, a head must do more than attend to that source: its write must follow the source token's identity. In matched prompts, replacing the value read at one source position while holding the attention pattern fixed shifts the selected heads' writes toward the replacement token and away from the original, more strongly than in control heads. Since the attention weights remain unchanged, this identity-sensitive effect comes from the source-dependent OV write rather than altered QK routing (Appendix~\ref{sec:app-fixed-attention}).

The sentence about computers supplies no matching capital-answer pattern, but copying need not depend on such a cue. In controlled prompts, we remove the earlier matching cue while keeping the candidate copied token, its position, and the final cue fixed. Copying weakens but does not disappear (Appendix~\ref{sec:app-cue}). We use parroting for this surviving cue-independent contribution. The same heads can also exploit an induction cue when one is present: induction and parroting describe behaviors, not mutually exclusive head populations.

\textbf{Copying can also combine several sources.} A prefix such as ``Springs compress under load'' contains several content-bearing words rather than a single candidate source. We compare the actual prefix-derived write with an attention-weighted mixture of source-token identity directions, then shuffle the pairing between tokens and attention weights while retaining the same tokens and weights. The correct pairing improves held-out reconstruction for a subset of the tested heads, supporting attention-weighted, multi-source copying (Appendix~\ref{sec:app-multisource}). We refer to this behavior as diffuse parroting. The mixture captures a distributed identity component, rather than everything the head writes.

Copying ``Computers'' does not necessarily mean carrying all the information associated with it. A pure identity vector can already align with related words, so we compare the actual write with a gain-matched identity-only reference. Association-related output remains beyond what this reference predicts. However, the before-and-after comparisons do not establish that the OV transformation preserves the measured association profile (Appendices~\ref{sec:app-association}--\ref{sec:app-association-edit}). A copying write can therefore contain more than token identity without establishing intact transfer of the source representation.

\textbf{Relational selectivity.} Making information about ``Computers'' available at END does not, by itself, explain why ``Berlin'' loses support. The query also indicates what kind of information matters: ``capital'' calls for a city. We examine middle-layer relational readers selected for attending to answer-bearing spans under golden context, then vary the relational frame independently of the candidate answer's type. For the Germany query, this means comparing a city such as ``Madrid'' inside and outside the capital-answer frame, as well as placing a non-city word inside that frame. Across our analyses a candidate of the expected answer type attracts strong attention even without the matching frame and all reader populations assign little attention to the tested unrelated filler (Appendix~\ref{sec:app-readers}).

\enlargethispage{24bp}
\begin{wrapfigure}{r}{0.68\textwidth}
\vspace{-\intextsep}
\centering
\includegraphics[width=\linewidth]{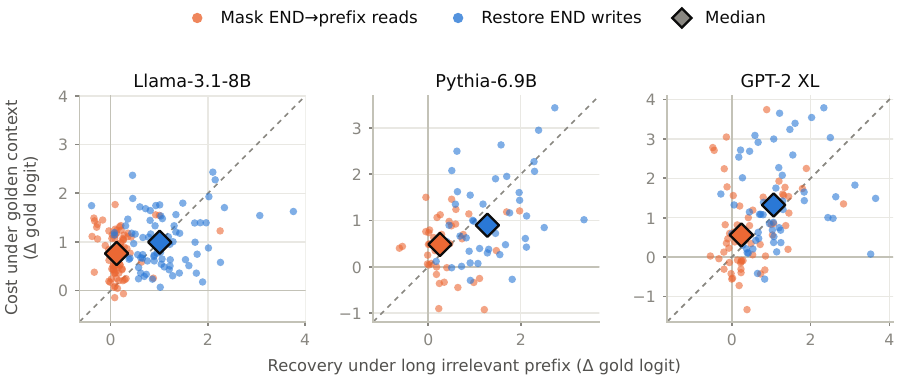}
\setlength{\abovecaptionskip}{2pt}
\caption{Per-fact recovery under the long irrelevant prefix versus cost under golden context, as gold-logit changes, for masking END$\to$prefix reads and restoring END writes; diamonds mark medians.}
\label{fig:middle-block}
\end{wrapfigure}
\textbf{Middle-block interventions.} Avoiding an irrelevant source is not the same as removing its influence from the answer computation (Figure~\ref{fig:pathway}). We therefore intervene on the broader middle block, rather than only the selected relational readers, replacing its attention heads' END-position writes with their context-free counterparts while retaining the prefixed input. Across the models and the short and long irrelevant-prefix conditions, this restores $41$--$90\%$ of the gold-logit loss. Masking the same heads' END-to-prefix attention instead increases the gold logit by at most $0.17$ (Appendix~\ref{sec:app-restoration}). Blocking these direct reads does not reproduce the recovery obtained by restoring the writes (Figure~\ref{fig:middle-block}). The restoration identifies an effective intervention site, but not a selective defence: it requires a context-free donor and also removes $45$--$87\%$ of the gold-logit benefit from golden context (Table~\ref{tab:app-golden}). The remaining question is whether a more specific intervention can reduce the influence of ``Computers are useful'' without sacrificing the useful evidence in ``Germany's capital city is Berlin.''

The preceding analyses explain how contextual information becomes available, but do not yet isolate a selective way to remove its harmful effects. Cancelling MLP writes also removes useful answer computation. The tested parroting interventions do not reliably separate unwanted prefix contributions from helpful-context use (Appendix~\ref{sec:app-copying-interventions}). The relational readers largely avoid unrelated filler, yet consequential changes remain in the broader middle block. Restoring that block's writes reduces interference, but requires a context-free donor and also removes useful support from golden context.

These analyses trace how contextual information is represented and propagated, but do not isolate a selective defence. We next identify individual heads associated with reduced answer support and test interventions on their contextual reads.
\wrapfill %

\section{Intrusion Heads: Localization and Intervention}
\label{sec:pathway-intrusion}

\textbf{Identifying intrusion heads.} A head's contextual read can change both what it adds from the prefix and what it contributes from the query. To identify the heads associated with lost answer support, we weight the reduction in each head's direct contribution to the correct-answer logit by its attention to the irrelevant prefix. This gives the intrusion score
\begin{equation}
  s_{\mathrm{int}}(h) = \frac{1}{N}\sum_{i=1}^{N} a_h^{(i,\mathrm{irr})} \max\!\left( 0,\, \mathrm{DLA}_h^{(i,\mathrm{nctx})}(g_i) - \mathrm{DLA}_h^{(i,\mathrm{irr})}(g_i) \right).
  \label{eq:intrusion-score}
\end{equation}

Here \(a_h\) sums attention from END to prefix positions, excluding BOS and the query, and the \(N\) discovery pairs give equal weight to short and long prefixes. We call the high-scoring population \textbf{intrusion heads} (\(I\)): their defining signature couples heads whose attention to the irrelevant prefix coincides with a reduction in their direct contribution to the correct-answer logit relative to the context-free run, rather than copying alone. 

\begin{figure}[!ht]
\centering
\begin{minipage}[t]{0.44\textwidth}
\vspace{0pt}
\centering
\includegraphics[scale=0.43]{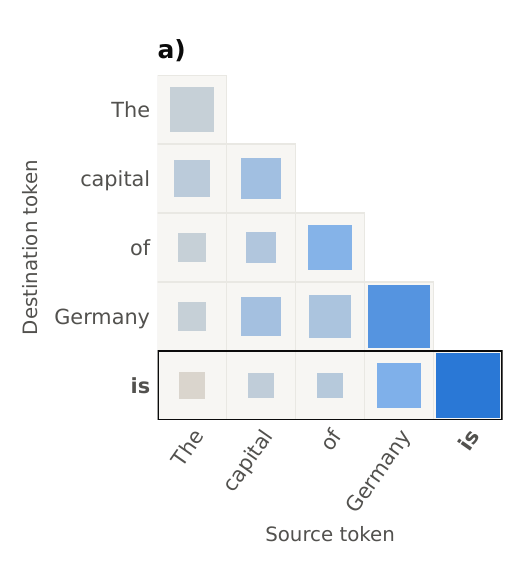}
\setlength{\abovecaptionskip}{4pt}
\caption{Net logit contribution (Berlin $-$ Computers) of each source token to each destination token in Llama-3.1-8B: (a) the bare query; (b) with the irrelevant prefix ``Computers are useful.''}
\label{fig:heatmaps}
\end{minipage}\hfill
\begin{minipage}[t]{0.54\textwidth}
\vspace{0pt}
\centering
\includegraphics[scale=0.43]{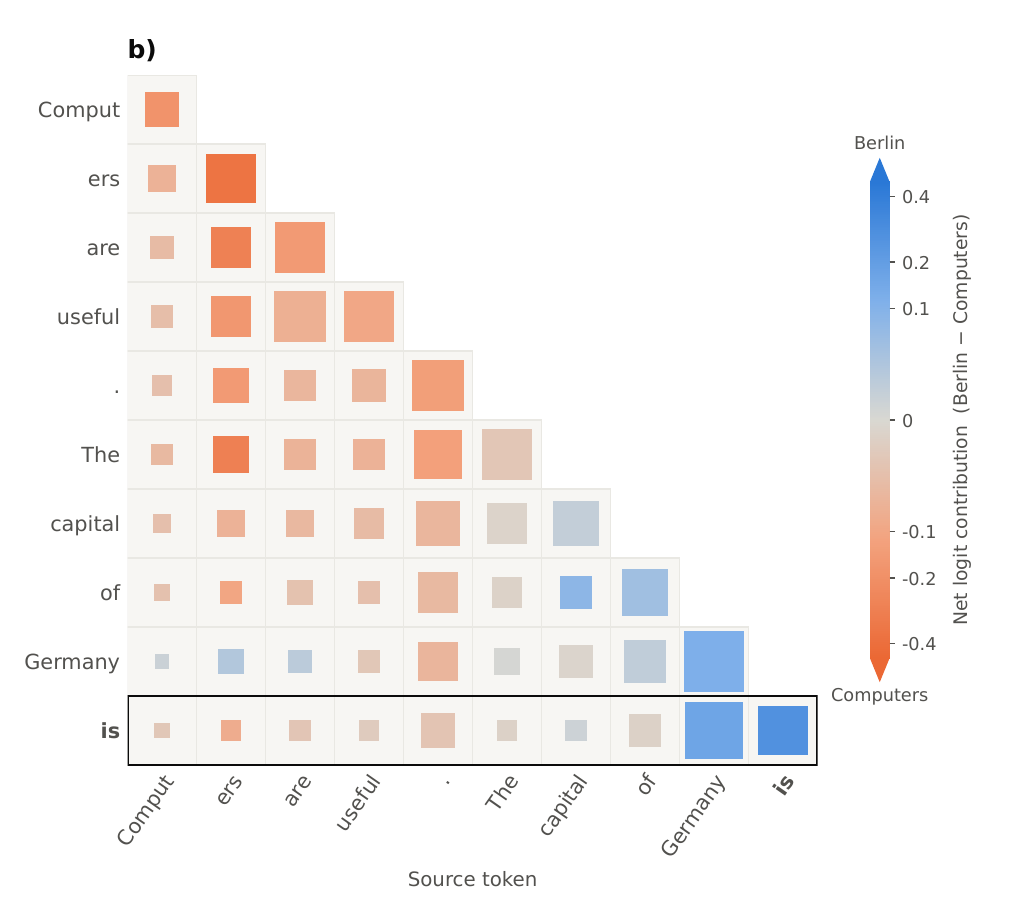}
\end{minipage}
\end{figure}

Having identified heads associated with reduced answer support, we inspect how their reads and writes change when irrelevant context is added. We select the intervention set \(I^\star\) using the intrusion score on discovery queries and keep it fixed for evaluation on held-out queries, testing whether its behavior extends beyond the examples used for selection (Appendix~\ref{sec:app-intrusion-populations}). Analyses of this set track three quantities: attention to the prefix and query, writes toward the correct answer, and distractor tokens, together with the associated words for each. In the context-free run, these heads attend to the query and BOS. Adding the irrelevant context shifts part of this attention toward the prefix. This change is accompanied by increased writing toward distractor tokens and their associated words, while their contributions to the correct answer weaken (Appendix~\ref{sec:app-intrusion}). Because distractor-directed writing does not enter the selection score, its increase constitutes an additional empirical finding. Figure~\ref{fig:heatmaps} illustrates the source-to-destination net logit contributions for the running example.

\enlargethispage{24bp}
\textbf{Prefix and non-prefix contributions.} To locate these changes, we split each write as \(o_h=P_h+U_h\), with \(P_h\) supplied by prefix positions and \(U_h\) by the remaining sources. Most of the increased distractor writing comes from \(P_h\), but this contribution has nearly zero mean direct gold logit attribution. The gold decrease instead comes from smaller contributions within \(U_h\), including subject, relation, and END positions (Appendix~\ref{sec:app-intrusion-sources}). This distinction is also evident in the matched-filler comparison: distractor promotion persists, but much of the gold-attribution difference shrinks (Figure~\ref{fig:app-intrusion-contrasts}). Thus, ``Computers'' is not simply read as a vector that opposes ``Berlin'': the prefix adds distractor-related content, but the loss of answer support comes from weaker query-derived contributions. This connects back to subject enrichment---information can remain available at the subject without contributing equally strongly to the answer.\looseness=-1

\textbf{Restoring answer-supporting writes.} The non-prefix contribution still supplies useful support for ``Berlin,'' so deleting the whole head would remove part of the answer computation along with the distractor-related write. We instead test whether correcting the context-dependent change restores that support. On held-out queries, replacing the END writes of the unfiltered population \(I\) with their context-free counterparts improves the gold logit more than matched-head controls, while the irrelevant prefix remains in the input. Interpolating as \(o_h^{(\lambda)}=o_h^{\mathrm{irr}}+\lambda(o_h^{\mathrm{nctx}}-o_h^{\mathrm{irr}})\) produces progressively stronger recovery as the correction moves from no replacement toward the full clean write (Appendix~\ref{sec:app-intrusion-restoration}). These interventions establish that the selected outputs are consequential, but still require a matched context-free pass. The source split suggests a finer alternative that does not: restrict the heads' prefix access while retaining the remaining sources.\looseness=-1

\enlargethispage{24bp}
\begin{wrapfigure}{l}{0.44\textwidth}
\vspace{-\intextsep}
\centering
\includegraphics[width=\linewidth]{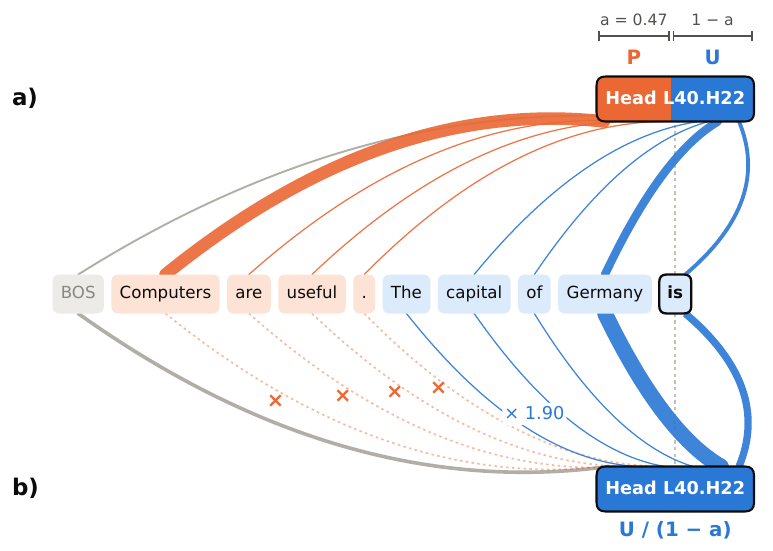}
\setlength{\abovecaptionskip}{2pt}
\caption{At fixed head inputs, masking combines prefix removal with amplification of the remaining contribution. It leaves the head in place and requires no clean donor.}
\label{fig:mask}
\end{wrapfigure}
\textbf{Targeted attention masking.} We implement this alternative by masking END-to-prefix attention in the fixed intervention set. Softmax redistributes the blocked attention across non-prefix positions, so the mask both removes the prefix write and strengthens the remaining reads (Figure~\ref{fig:mask}). At fixed head inputs,
\begin{equation}
  o_h^{\mathrm{mask}}-o_h = -P_h+\frac{a_h}{1-a_h}U_h, \qquad a_h<1.
  \label{eq:mask-decomposition}
\end{equation}
We separate these operations by removing \(P_h\) without redistributing its attention, amplifying \(U_h\) while retaining \(P_h\), and applying both changes together. Prefix removal supplies most of the distractor reduction, whereas amplification increases the edited write's direct gold support. Where mask improves the final prediction, its benefit can arise from both removal and amplification; applying the same gain to the complete heads' write does not reproduce that benefit (Appendix~\ref{sec:app-four-writes}). The full mask can therefore improve gold probability and lower distractor support on the same held-out queries without deleting the heads or consulting a clean run (Appendix~\ref{sec:app-two-sided-mask}).

\textbf{Direct versus copier-mediated access.} Does this contextual read depend on information first copied into END? We replace the measured copiers' contribution to the intrusion heads' END query inputs with its context-free counterpart. This barely changes their attention and produces no substantial beneficial recovery of prediction loss. A complementary intervention targets the prefix itself: its keys and values are replaced with those from corresponding token positions in a filler run, leaving END unpatched. In settings where masking improves prediction, this intervention recovers part of that improvement. The consequential access can therefore be direct: the tested copier-to-query route does not account for the effect of changing what these heads read from the prefix (Appendix~\ref{sec:app-intrusion-route}).

\textbf{Comparison with entrainment and retrieval heads.} The distractor-directed signal also invites comparison with contextual entrainment, in which heads just promote the tokens already present in the prefix \citep{niu2025llama}. This behavioral overlap does not establish that the same head population is involved. We therefore compare the independently ranked intrusion heads with the entrainment set. It contains none of the twenty highest intrusion-scored heads (Appendix~\ref{sec:app-entrainment}). The overlap in behavior thus does not amount to rediscovering that entrainment population.

\begin{wrapfigure}{l}{0.42\textwidth}
\vspace{-\intextsep}
\centering
\includegraphics[width=\linewidth]{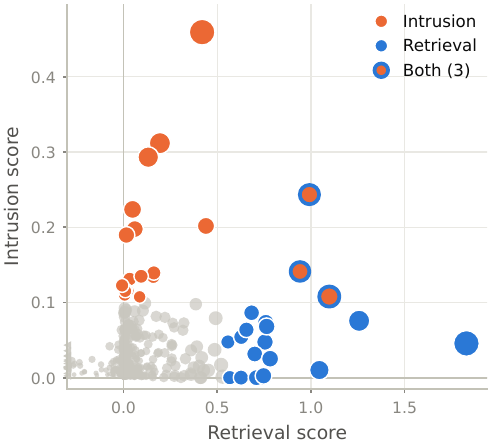}
\setlength{\abovecaptionskip}{2pt}
\caption{Retrieval score versus intrusion score for each head in GPT-2 XL.}
\label{fig:lshape}
\end{wrapfigure}
Because the intrusion score identifies heads through reduced gold support under irrelevant context, we must also check whether masking them preserves answer support when the prefix is helpful. In our running example, reducing interference from ``Computers are useful'' should not prevent the model from using ``Germany's capital city is Berlin.'' We therefore compare intrusion heads with contextual retrieval heads \citep{wu2024retrieval}, scored by their attention to the answer's position in the golden prefix and their gold-directed write. Comparing the independently ranked populations before any exclusions shows that their top-scoring heads are different, although some overlap remains (Figure~\ref{fig:lshape}; Appendix~\ref{sec:app-contextual-retrieval}). With golden context, this targeted mask retains the benefit supplied by the evidence (Appendices~\ref{sec:app-two-sided-mask} and~\ref{sec:app-contextual-retrieval}). Thus, restricting contextual access in the selected intrusion heads can reduce interference while largely preserving the model's ability to use helpful context, rather than achieving improvement by blocking context everywhere.
\wrapfill

\section{Downstream Evaluation}
\label{sec:downstream}

The pathway identified in \S\ref{sec:pathway} suggests a practical intervention with two requirements. The targeted intrusion heads must be separable from the heads that integrate evidential context, as \S\ref{sec:pathway-intrusion} establishes. They must also remain useful intervention targets beyond the task on which they were identified, so that a model-specific set can be identified once on a simple benchmark and reused. We test this second requirement through independent head re-identification and fixed-set intervention on four downstream tasks.

\paragraph{Evaluation setting.} We evaluate four base models: Llama-3.1-8B, Mistral-7B-v0.1, Qwen2.5-7B, and Pythia-6.9B and use the same four benchmarks to assess cross-task head stability and downstream intervention effectiveness. Each pairs an original task input with controlled irrelevant context; the clean condition retains the original question and any supporting evidence. GSM-DC~\citep{yang2025gsmdc} injects \(k\) type-consistent but irrelevant arithmetic nodes into GSM8K dependency graphs. GSM-IC~\citep{shi2023distracted} inserts one irrelevant sentence into the original problem. HotpotQA~\citep{yang2018hotpotqa} retains the two supporting passages and adds \(k\) non-supporting Wikipedia paragraphs. PubMedQA~\citep{jin2019pubmedqa} retains the supporting abstract and appends \(k\) irrelevant abstracts. These settings include in-domain distractors, extending the evaluation beyond the unrelated prefixes used for head identification. We report recovery as the percentage of distractor-induced accuracy loss restored by an intervention; task-specific metrics, sample sizes, and per-\(k\) results are provided in Appendix~\ref{sec:appendix-downstream}.

\paragraph{Cross-task stability.} If intrusion reflects a shared context-processing operation rather than a task-specific subroutine, different tasks should recruit overlapping intrusion-head populations, consistent with findings that head-level roles are reused across tasks \citep{wu2024retrieval,merullo2024circuit}. We test this prediction within each of the four base models by independently recomputing the intrusion score in Eq.~\ref{eq:intrusion-score} on factual recall, binding/retrieval (IOI), and the four downstream tasks, and comparing the resulting rankings and selected head sets. As a within-task reference, independent rankings from $1{,}000$ random halves of the factual-recall data yield Spearman correlations of $0.97$--$0.99$ across models. Table~\ref{tab:cross-task} summarizes cross-task overlap among the ten-head sets.

\begin{figure}[t]
\centering
\includegraphics[width=\textwidth]{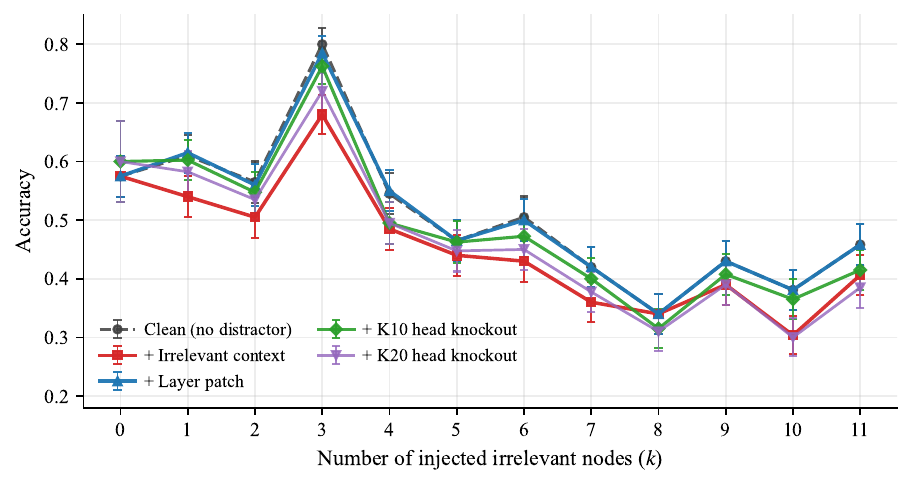}
\caption{GSM-DC across $k$: layer patching closes the gap at every $k$; K10 recovers a large fraction with $\sim$10 heads. Per-model breakdowns in Appendix~\ref{sec:appendix-downstream}.}
\label{fig:downstream}
\end{figure}

\begin{wraptable}{r}{0.40\textwidth}
\vspace{-\intextsep}
\centering
\footnotesize
\setlength{\tabcolsep}{3pt}
\renewcommand{\arraystretch}{1.05}
\setlength{\abovecaptionskip}{4pt}
\begin{tabular}{@{}lrrr@{}}
\toprule
\textbf{Task} & \textbf{+Patch} & \textbf{+K10} & \textbf{+K20} \\
\midrule
GSM-DC      & $96.7\%$  & $57.3\%$ & $28.8\%$ \\
GSM-IC      & $108.8\%$ & $58.5\%$ & $27.8\%$ \\
HotpotQA    & $85.5\%$  & $48.5\%$ & $29.5\%$ \\
PubMedQA    & $99.3\%$  & $63.3\%$ & $35.9\%$ \\
\midrule
\textbf{mean} & $\mathbf{97.6\%}$ & $\mathbf{56.9\%}$ & $\mathbf{30.5\%}$ \\
\bottomrule
\end{tabular}
\caption{Per-task recovery of the distractor-induced drop (7B baselines): +Patch layer residual patch, +K10 clean intrusion-head knockout ($\sim$10 heads), +K20 top-20 incl. dual-use. (>=100\% = above baseline)}
\label{tab:recovery}
\vspace{-\intextsep}
\end{wraptable}
\paragraph{Fixed-head attention masking.} We first test the practical intervention suggested by \S\ref{sec:pathway-intrusion}. For the frozen, retrieval-disjoint intrusion-head set, we mask the pre-softmax attention scores from END to context positions, redistributing attention to non-context positions while leaving all other heads unmodified. This suppresses the selected attention edges rather than deleting the heads or their weights. The intervention recovers $48.5$--$63.3\%$ of the distractor-induced accuracy loss across the four tasks (Table~\ref{tab:recovery}), without a matched clean forward pass. The GSM-DC sweep further shows that its benefit is not restricted to a single distractor count (Figure~\ref{fig:downstream}). A broader knockout that includes dual-use heads recovers less (Appendix~\ref{sec:appendix-downstream}), consistent with the importance of preserving useful context readers rather than indiscriminately suppressing context attention.

\paragraph{Oracle residual patching.} To compare this sparse intervention with a broader oracle correction, we replace the entire END-position residual at the model-specific intrusion-band layer with its counterpart from the matched clean input. The substitution is applied per generation step, while other token positions are not directly patched. This oracle recovers $97.6\%$ of the distractor-induced loss on average, ranging from $85.5\%$ on HotpotQA to $108.8\%$ on GSM-IC (Table~\ref{tab:recovery}). Recovery above $100\%$ means that the patched model exceeds the unmodified clean baseline. One plausible explanation is that replacing the accumulated contamination at END, while retaining potentially useful contextual representations at other positions, allows subsequent layers to combine a restored answer state with those representations under reduced interference. Because the remaining contextual states still come from the distracted run, this intervention need not merely reproduce the clean computation. The patch therefore provides an oracle reference for how much loss can be reversed at the localized residual state, whereas knockout measures what can be recovered through a fixed, sparse intervention.

Together, these results turn the mechanistic localization into a reusable intervention: a fixed mask affecting approximately $1\%$ of attention heads recovers $56.9\%$ of distractor-induced accuracy loss on average across four tasks. It requires neither downstream head reselection nor a matched clean forward pass at inference. Oracle patching, which does require that clean pass, marks the ceiling: nearly all of the loss is reversible at the localized state, and knockout is the share of that ceiling reachable at deployment.

\section{Conclusion}
\label{sec:conclusion}

We provide a mechanistic account of how non-evidential context weakens answer support, distinguish this loss from contextual token promotion alone, and use that understanding to guide a targeted intervention. We find answer support can weaken despite preserved subject enrichment and selective reading. Broad write restoration recovers support but requires a context-free donor and also removes golden-context benefits. We then localize a recoverable component of the interference to a sparse set of intrusion heads. These heads are largely distinct from the tested entrainment and golden-context retrieval populations. This localization guides a selective intervention that restricts contextual reads in the identified intrusion heads.

The resulting training-free mask targets roughly $1\%$ of all attention heads. Applied unchanged across four downstream tasks, it recovers $57\%$ of distractor-induced accuracy loss on average while largely preserving golden-context benefits. Broader oracle residual patching recovers $97.6\%$ on average but required a matched clean pass. Retrieval-augmented generation, tool use, agentic loops, and long-context settings motivates this problem: helpful evidence and non-evidential material can arrive through the same context channel. A substantial part of contextual interference can be localized and mitigated within the computation itself, without retraining the model or broadly disabling its access to useful evidence.

\section*{Limitations}

Our findings concern the evaluated open-weight decoder-only base models, with four 7B-scale models forming the main evaluation and GPT-2 XL included in circuit diagnostics. Broader applicability to substantially larger models, mixture-of-experts architectures, encoder--decoder models, and multimodal systems requires separate validation. The circuit analyses use factual recall and IOI binding/retrieval, while downstream evaluation covers four English benchmarks with controlled distractors, including in-domain material. These settings do not establish robustness for code, multilingual prompts, multi-turn dialogue, mathematics beyond the evaluated GSM tasks, or adversarial and jailbreak-style inputs.

Our measurements concern selected components, not complete representations or their training origins; causal evidence comes from interventions rather than logit attribution alone. The fixed mask requires internal attention access, model-specific head selection from paired inputs, and specified context positions, but does not determine relevance. Recovery remains partial, and helpful-context preservation is empirical rather than guaranteed. Oracle patching requires matched clean activations and provides a diagnostic reference, not a deployable solution or formal upper bound. Comparisons with softer and learned interventions remain future work. Our experiments do not establish a minimal head set or an optimal trade-off between robustness and helpful-context preservation.

\vspace{37.5pt}
\bibliography{custom}

\begin{thebibliography}{37}
\providecommand{\natexlab}[1]{#1}
\providecommand{\url}[1]{\texttt{#1}}
\expandafter\ifx\csname urlstyle\endcsname\relax
  \providecommand{\doi}[1]{doi: #1}\else
  \providecommand{\doi}{doi: \begingroup \urlstyle{rm}\Url}\fi

\bibitem[Asai et~al.(2024)Asai, Wu, Wang, Sil, and Hajishirzi]{asai2024selfrag}
Akari Asai, Zeqiu Wu, Yizhong Wang, Avirup Sil, and Hannaneh Hajishirzi.
\newblock {Self-RAG}: Learning to retrieve, generate, and critique through
  self-reflection.
\newblock In \emph{International Conference on Learning Representations
  (ICLR)}, 2024.

\bibitem[Biderman et~al.(2023)Biderman, Schoelkopf, Anthony,
  et~al.]{biderman2023pythia}
Stella Biderman, Hailey Schoelkopf, Quentin~Gregory Anthony, et~al.
\newblock {Pythia}: A suite for analyzing large language models across training
  and scaling.
\newblock In \emph{Proceedings of the 40th International Conference on Machine
  Learning}, volume 202 of \emph{Proceedings of Machine Learning Research},
  pp.\  2397--2430. PMLR, 2023.
\newblock URL \url{https://proceedings.mlr.press/v202/biderman23a.html}.

\bibitem[Chowdhury(2026)]{chowdhury2026circuits}
Borun~D. Chowdhury.
\newblock Lost in the middle at birth: An exact theory of transformer position
  bias, 2026.
\newblock URL \url{https://arxiv.org/abs/2603.10123}.

\bibitem[Cuconasu et~al.(2024)Cuconasu, Trappolini, Siciliano, Filice,
  Campagnano, Maarek, Tonellotto, and Silvestri]{cuconasu2024noise}
Florin Cuconasu, Giovanni Trappolini, Federico Siciliano, Simone Filice, Cesare
  Campagnano, Yoelle Maarek, Nicola Tonellotto, and Fabrizio Silvestri.
\newblock The power of noise: Redefining retrieval for {RAG} systems.
\newblock In \emph{Proceedings of the 47th International ACM SIGIR Conference
  on Research and Development in Information Retrieval (SIGIR)}, pp.\
  719--729, 2024.

\bibitem[Fang et~al.(2024)Fang, Bai, Ni, Yang, Chen, and Xu]{fang2024noise}
Feiteng Fang, Yuelin Bai, Shiwen Ni, Min Yang, Xiaojun Chen, and Ruifeng Xu.
\newblock Enhancing noise robustness of retrieval-augmented language models
  with adaptive adversarial training.
\newblock In \emph{Proceedings of the 62nd Annual Meeting of the Association
  for Computational Linguistics (ACL), Long Papers}, pp.\  10028--10039, 2024.

\bibitem[Feng \& Steinhardt(2024)Feng and Steinhardt]{feng2024binding}
Jiahai Feng and Jacob Steinhardt.
\newblock How do language models bind entities in context?
\newblock In \emph{International Conference on Learning Representations
  (ICLR)}, 2024.

\bibitem[Geva et~al.(2023)Geva, Bastings, Filippova, and
  Globerson]{geva2023dissecting}
Mor Geva, Jasmijn Bastings, Katja Filippova, and Amir Globerson.
\newblock Dissecting recall of factual associations in auto-regressive language
  models.
\newblock In \emph{Proceedings of the 2023 Conference on Empirical Methods in
  Natural Language Processing (EMNLP)}, pp.\  12216--12235, 2023.

\bibitem[Goldowsky-Dill et~al.(2023)Goldowsky-Dill, MacLeod, Sato, and
  Arora]{goldowsky2023localizing}
Nicholas Goldowsky-Dill, Chris MacLeod, Lucas Sato, and Aryaman Arora.
\newblock Localizing model behavior with path patching.
\newblock \emph{arXiv preprint arXiv:2304.05969}, 2023.

\bibitem[Grattafiori et~al.(2024)Grattafiori, Dubey, Jauhri,
  et~al.]{grattafiori2024llama3}
Aaron Grattafiori, Abhimanyu Dubey, Abhinav Jauhri, et~al.
\newblock The {Llama 3} herd of models.
\newblock arXiv preprint arXiv:2407.21783, 2024.
\newblock URL \url{https://arxiv.org/abs/2407.21783}.

\bibitem[Hendel et~al.(2023)Hendel, Geva, and Globerson]{hendel2023task}
Roee Hendel, Mor Geva, and Amir Globerson.
\newblock In-context learning creates task vectors.
\newblock In \emph{Findings of the Association for Computational Linguistics:
  EMNLP 2023}, pp.\  9318--9333, 2023.

\bibitem[Hernandez et~al.(2024)Hernandez, Sharma, Haklay, Meng, Wattenberg,
  Andreas, Belinkov, and Bau]{hernandez2023lre}
Evan Hernandez, Arnab~Sen Sharma, Tal Haklay, Kevin Meng, Martin Wattenberg,
  Jacob Andreas, Yonatan Belinkov, and David Bau.
\newblock Linearity of relation decoding in transformer language models.
\newblock In \emph{International Conference on Learning Representations
  (ICLR)}, 2024.

\bibitem[Jiang et~al.(2023)Jiang, Sablayrolles, Mensch,
  et~al.]{jiang2023mistral}
Albert~Q. Jiang, Alexandre Sablayrolles, Arthur Mensch, et~al.
\newblock {Mistral 7B}.
\newblock arXiv preprint arXiv:2310.06825, 2023.
\newblock URL \url{https://arxiv.org/abs/2310.06825}.

\bibitem[Jin et~al.(2019)Jin, Dhingra, Liu, Cohen, and Lu]{jin2019pubmedqa}
Qiao Jin, Bhuwan Dhingra, Zhengping Liu, William~W. Cohen, and Xinghua Lu.
\newblock {PubMedQA}: A dataset for biomedical research question answering.
\newblock In \emph{Proceedings of the 2019 Conference on Empirical Methods in
  Natural Language Processing and the 9th International Joint Conference on
  Natural Language Processing (EMNLP-IJCNLP)}, pp.\  2567--2577, 2019.

\bibitem[Lewis et~al.(2020)Lewis, Perez, Piktus, Petroni, Karpukhin, Goyal,
  K{\"u}ttler, Lewis, Yih, Rockt{\"a}schel, Riedel, and Kiela]{lewis2020rag}
Patrick Lewis, Ethan Perez, Aleksandra Piktus, Fabio Petroni, Vladimir
  Karpukhin, Naman Goyal, Heinrich K{\"u}ttler, Mike Lewis, Wen-tau Yih, Tim
  Rockt{\"a}schel, Sebastian Riedel, and Douwe Kiela.
\newblock Retrieval-augmented generation for knowledge-intensive {NLP} tasks.
\newblock In \emph{Advances in Neural Information Processing Systems 33
  (NeurIPS)}, pp.\  9459--9474, 2020.

\bibitem[Longpre et~al.(2021)Longpre, Perisetla, Chen, Ramesh, DuBois, and
  Singh]{longpre2021entity}
Shayne Longpre, Kartik Perisetla, Anthony Chen, Nikhil Ramesh, Chris DuBois,
  and Sameer Singh.
\newblock Entity-based knowledge conflicts in question answering.
\newblock In \emph{Proceedings of the 2021 Conference on Empirical Methods in
  Natural Language Processing (EMNLP)}, pp.\  7052--7063, 2021.

\bibitem[Mallen et~al.(2023)Mallen, Asai, Zhong, Das, Khashabi, and
  Hajishirzi]{mallen2023trust}
Alex Mallen, Akari Asai, Victor Zhong, Rajarshi Das, Daniel Khashabi, and
  Hannaneh Hajishirzi.
\newblock When not to trust language models: Investigating effectiveness of
  parametric and non-parametric memories.
\newblock In \emph{Proceedings of the 61st Annual Meeting of the Association
  for Computational Linguistics (ACL), Long Papers}, pp.\  9802--9822, 2023.

\bibitem[McDougall et~al.(2023)McDougall, Conmy, Rushing, McGrath, and
  Nanda]{mcdougall2023copy}
Callum McDougall, Arthur Conmy, Cody Rushing, Thomas McGrath, and Neel Nanda.
\newblock Copy suppression: Comprehensively understanding an attention head.
\newblock \emph{arXiv preprint arXiv:2310.04625}, 2023.

\bibitem[Meng et~al.(2022)Meng, Bau, Andonian, and Belinkov]{meng2022rome}
Kevin Meng, David Bau, Alex Andonian, and Yonatan Belinkov.
\newblock Locating and editing factual associations in {GPT}.
\newblock In \emph{Advances in Neural Information Processing Systems 35
  (NeurIPS)}, pp.\  17359--17372, 2022.

\bibitem[Merity et~al.(2016)Merity, Xiong, Bradbury, and Socher]{wikitext103}
Stephen Merity, Caiming Xiong, James Bradbury, and Richard Socher.
\newblock Pointer sentinel mixture models, 2016.

\bibitem[Merullo et~al.(2024)Merullo, Eickhoff, and
  Pavlick]{merullo2024circuit}
Jack Merullo, Carsten Eickhoff, and Ellie Pavlick.
\newblock Circuit component reuse across tasks in transformer language models.
\newblock In \emph{International Conference on Learning Representations
  (ICLR)}, 2024.

\bibitem[Nanda \& Bloom(2023)Nanda and Bloom]{nanda2023tlens}
Neel Nanda and Joseph Bloom.
\newblock {TransformerLens}: A library for mechanistic interpretability of
  generative language models.
\newblock \url{https://github.com/neelnanda-io/TransformerLens}, 2023.

\bibitem[Niu et~al.(2025)Niu, Yuan, Wang, Saghir, and Abdi]{niu2025llama}
Jingcheng Niu, Xingdi Yuan, Tong Wang, Hamidreza Saghir, and Amir~H. Abdi.
\newblock Llama see, llama do: A mechanistic perspective on contextual
  entrainment and distraction in {LLMs}.
\newblock In \emph{Proceedings of the 63rd Annual Meeting of the Association
  for Computational Linguistics (ACL), Long Papers}, pp.\  16218--16239, 2025.
\newblock \doi{10.18653/v1/2025.acl-long.791}.
\newblock URL \url{https://aclanthology.org/2025.acl-long.791/}.

\bibitem[Olsson et~al.(2022)Olsson, Elhage, Nanda, Joseph, DasSarma, Henighan,
  Mann, Askell, Bai, Chen, Conerly, Drain, Ganguli, Hatfield-Dodds, Hernandez,
  Johnston, Jones, Kernion, Lovitt, Ndousse, Amodei, Brown, Clark, Kaplan,
  McCandlish, and Olah]{olsson2022induction}
Catherine Olsson, Nelson Elhage, Neel Nanda, Nicholas Joseph, Nova DasSarma,
  Tom Henighan, Ben Mann, Amanda Askell, Yuntao Bai, Anna Chen, Tom Conerly,
  Dawn Drain, Deep Ganguli, Zac Hatfield-Dodds, Danny Hernandez, Scott
  Johnston, Andy Jones, Jackson Kernion, Liane Lovitt, Kamal Ndousse, Dario
  Amodei, Tom Brown, Jack Clark, Jared Kaplan, Sam McCandlish, and Chris Olah.
\newblock In-context learning and induction heads.
\newblock \emph{Transformer Circuits Thread}, 2022.

\bibitem[Pennington et~al.(2014)Pennington, Socher, and
  Manning]{pennington2014glove}
Jeffrey Pennington, Richard Socher, and Christopher~D. Manning.
\newblock {GloVe}: Global vectors for word representation.
\newblock In \emph{Proceedings of the 2014 Conference on Empirical Methods in
  Natural Language Processing (EMNLP)}, pp.\  1532--1543, 2014.

\bibitem[{Qwen Team}(2024)]{yang2024qwen25}
{Qwen Team}.
\newblock {Qwen2.5} technical report.
\newblock arXiv preprint arXiv:2412.15115, 2024.
\newblock URL \url{https://arxiv.org/abs/2412.15115}.

\bibitem[Radford et~al.(2019)Radford, Wu, Child, Luan, Amodei, and
  Sutskever]{radford2019language}
Alec Radford, Jeffrey Wu, Rewon Child, David Luan, Dario Amodei, and Ilya
  Sutskever.
\newblock Language models are unsupervised multitask learners.
\newblock Technical report, OpenAI, 2019.
\newblock URL
  \url{https://cdn.openai.com/better-language-models/language_models_are_unsupervised_multitask_learners.pdf}.

\bibitem[Schick et~al.(2023)Schick, Dwivedi-Yu, Dess{\`\i}, Raileanu, Lomeli,
  Zettlemoyer, Cancedda, and Scialom]{schick2023toolformer}
Timo Schick, Jane Dwivedi-Yu, Roberto Dess{\`\i}, Roberta Raileanu, Maria
  Lomeli, Luke Zettlemoyer, Nicola Cancedda, and Thomas Scialom.
\newblock Toolformer: Language models can teach themselves to use tools.
\newblock In \emph{Advances in Neural Information Processing Systems 36
  (NeurIPS)}, 2023.

\bibitem[Shi et~al.(2023)Shi, Chen, Misra, Scales, Dohan, Chi, Sch{\"a}rli, and
  Zhou]{shi2023distracted}
Freda Shi, Xinyun Chen, Kanishka Misra, Nathan Scales, David Dohan, Ed~H. Chi,
  Nathanael Sch{\"a}rli, and Denny Zhou.
\newblock Large language models can be easily distracted by irrelevant context.
\newblock In \emph{Proceedings of the 40th International Conference on Machine
  Learning (ICML)}, pp.\  31210--31227, 2023.

\bibitem[Wang et~al.(2023)Wang, Variengien, Conmy, Shlegeris, and
  Steinhardt]{wang2023ioi}
Kevin Wang, Alexandre Variengien, Arthur Conmy, Buck Shlegeris, and Jacob
  Steinhardt.
\newblock Interpretability in the wild: A circuit for indirect object
  identification in {GPT-2} small.
\newblock In \emph{International Conference on Learning Representations
  (ICLR)}, 2023.

\bibitem[Wu et~al.(2025{\natexlab{a}})Wu, Zhang, Che, Feng, Shao, and
  Tao]{wu2025pandora}
Jinyang Wu, Shuai Zhang, Feihu Che, Mingkuan Feng, Pengpeng Shao, and Jianhua
  Tao.
\newblock Pandora's box or aladdin's lamp: A comprehensive analysis revealing
  the role of {RAG} noise in large language models.
\newblock In \emph{Proceedings of the 63rd Annual Meeting of the Association
  for Computational Linguistics (ACL), Long Papers}, pp.\  5019--5039,
  2025{\natexlab{a}}.
\newblock \doi{10.18653/v1/2025.acl-long.250}.
\newblock URL \url{https://aclanthology.org/2025.acl-long.250/}.

\bibitem[Wu et~al.(2024)Wu, Xie, Chen, Zhu, Zhang, and Xiao]{wu2024easily}
Siye Wu, Jian Xie, Jiangjie Chen, Tinghui Zhu, Kai Zhang, and Yanghua Xiao.
\newblock How easily do irrelevant inputs skew the responses of large language
  models?
\newblock In \emph{First Conference on Language Modeling (COLM)}, 2024.

\bibitem[Wu et~al.(2025{\natexlab{b}})Wu, Wang, Xiao, Peng, and
  Fu]{wu2024retrieval}
Wenhao Wu, Yizhong Wang, Guangxuan Xiao, Hao Peng, and Yao Fu.
\newblock Retrieval head mechanistically explains long-context factuality.
\newblock In \emph{International Conference on Learning Representations
  (ICLR)}, 2025{\natexlab{b}}.

\bibitem[Yang et~al.(2025)Yang, Huang, Zhang, Surdeanu, Wang, and
  Pan]{yang2025gsmdc}
Minglai Yang, Ethan Huang, Liang Zhang, Mihai Surdeanu, William~Yang Wang, and
  Liangming Pan.
\newblock How is {LLM} reasoning distracted by irrelevant context? an analysis
  using a controlled benchmark.
\newblock In \emph{Proceedings of the 2025 Conference on Empirical Methods in
  Natural Language Processing}, pp.\  13329--13347. Association for
  Computational Linguistics, 2025.
\newblock \doi{10.18653/v1/2025.emnlp-main.674}.
\newblock URL \url{https://aclanthology.org/2025.emnlp-main.674/}.

\bibitem[Yang et~al.(2018)Yang, Qi, Zhang, Bengio, Cohen, Salakhutdinov, and
  Manning]{yang2018hotpotqa}
Zhilin Yang, Peng Qi, Saizheng Zhang, Yoshua Bengio, William~W. Cohen, Ruslan
  Salakhutdinov, and Christopher~D. Manning.
\newblock {HotpotQA}: A dataset for diverse, explainable multi-hop question
  answering.
\newblock In \emph{Proceedings of the 2018 Conference on Empirical Methods in
  Natural Language Processing (EMNLP)}, pp.\  2369--2380, 2018.

\bibitem[Yao et~al.(2023)Yao, Zhao, Yu, Du, Shafran, Narasimhan, and
  Cao]{yao2023react}
Shunyu Yao, Jeffrey Zhao, Dian Yu, Nan Du, Izhak Shafran, Karthik Narasimhan,
  and Yuan Cao.
\newblock {ReAct}: Synergizing reasoning and acting in language models.
\newblock In \emph{International Conference on Learning Representations
  (ICLR)}, 2023.

\bibitem[Yoran et~al.(2024)Yoran, Wolfson, Ram, and Berant]{yoran2024making}
Ori Yoran, Tomer Wolfson, Ori Ram, and Jonathan Berant.
\newblock Making retrieval-augmented language models robust to irrelevant
  context.
\newblock In \emph{International Conference on Learning Representations
  (ICLR)}, 2024.

\bibitem[Yu et~al.(2024)Yu, Zhang, Pan, Cao, Ma, Li, Wang, and
  Yu]{yu2024chainofnote}
Wenhao Yu, Hongming Zhang, Xiaoman Pan, Peixin Cao, Kaixin Ma, Jian Li, Hongwei
  Wang, and Dong Yu.
\newblock Chain-of-note: Enhancing robustness in retrieval-augmented language
  models.
\newblock In \emph{Proceedings of the 2024 Conference on Empirical Methods in
  Natural Language Processing (EMNLP)}, pp.\  14672--14685, 2024.

\end{thebibliography}
\bibliographystyle{preprint}
\appendix
\counterwithin{table}{section}
\counterwithin{figure}{section}
\counterwithin{equation}{section}
\renewcommand{\floatpagefraction}{0.92}
\renewcommand{\textfraction}{0.05}
\renewcommand{\bottomfraction}{0.6}
\setcounter{topnumber}{4}
\setcounter{bottomnumber}{2}
\setcounter{totalnumber}{6}
\makeatletter
\setlength{\@fptop}{0pt}
\setlength{\@fpsep}{12pt plus 2pt}
\setlength{\@fpbot}{0pt plus 1fil}
\makeatother

\section{Setup}
\label{app:data}

\subsection{Factual-recall benchmark: relation composition}
\label{app:data:lre}

The factual-recall benchmark draws on relations from LRE~\citep{hernandez2023lre}, selected to span the three canonical knowledge regimes that the factual-recall literature has characterized at the circuit level. Factual associations --- the regime ROME~\citep{meng2022rome} and the Geva--Bastings subject-enrichment account \citep{geva2023dissecting} localize to mid-layer subject MLPs --- are represented by relations such as \texttt{country\_capital\_city}, \texttt{country\_language}, and \texttt{person\_plays\_instrument}. Categorical relations, the regime that produced the Hernandez \textit{et al.}\ LRE benchmark itself, are represented by \texttt{object\_superclass} and its kin. Geographic composition --- the regime where binding-ID mechanisms~\citep{feng2024binding} have been characterized --- is represented by \texttt{city\_in\_country} and \texttt{landmark\_in\_country}.

\paragraph{Relation distribution.}
The qualitative conclusions of \S\ref{sec:pathway} hold within every relation taken individually.

\subsection{Binding/retrieval benchmark: IOI prompts}
\label{app:data:ioi}

To complement the factual-recall task family, we include the Indirect Object Identification (IOI) benchmark of \citet{wang2023ioi}, which probes a structurally distinct circuit: binding entity tokens to syntactic roles and retrieving the correct role-bound entity at the END position. We follow Wang et al.'s prompt-generation procedure exactly --- IOI prompts are produced by combinatorial sampling over four pools:
\begin{itemize}\setlength{\itemsep}{1pt}
\item \textbf{Name pool} (single-token first names from the Wang-2023 release).
\item \textbf{Place pool} (e.g., \textit{store}, \textit{school}, \textit{park}, \textit{garden}, \textit{restaurant}, \textit{hospital}).
\item \textbf{Object pool} (e.g., \textit{drink}, \textit{ball}, \textit{book}, \textit{ring}, \textit{kiss}, \textit{pen}).
\item \textbf{Template inventory} of syntactic frames, e.g.\ ``When $[$Name$_1]$ and $[$Name$_2]$ went to the $[$place$]$, $[$Name$_i]$ gave a $[$object$]$ to \_\_\_'' and analogous ``At/Then/After'' variants.
\end{itemize}
Each prompt has the canonical IOI structure $[\text{Name}_A \dots \text{Name}_B \dots \text{Name}_S$\,gave/received$\,\dots]$, with the gold answer $a^\star$ being the indirect object (the name in the IO syntactic role). The dataset is balanced across \textbf{ABBA} and \textbf{BABA} surface orderings (the IO appears in the first or second position with equal frequency), exactly as in \citet{wang2023ioi}. Each prompt is instantiated under the same seven-way prefix split as the factual-recall samples (\texttt{nctx}, \texttt{golden}, \texttt{related}, \texttt{counterfactual}, \texttt{short\_irr}, \texttt{long\_irr}, \texttt{random}), so the two task families share identical paired-comparison machinery and downstream analysis pipeline.

\subsection{Representative data samples}
\label{app:data:samples}

We provide concrete examples to make the seven-way prefix split unambiguous. The running query is \emph{``The capital of Germany is \_\_\_''} ($a^\star = $ Berlin) for factual recall and \emph{``When Mary and John went to the store, John gave a drink to \_\_\_''} ($a^\star = $ Mary) for IOI.

\paragraph{Factual recall (LRE \texttt{country\_capital\_city}).}
\begin{itemize}\setlength{\itemsep}{1pt}
\item \texttt{nctx}: \texttt{"The capital of Germany is"}
\item \texttt{golden}: \texttt{"Germany's capital city is Berlin. The capital of Germany is"}
\item \texttt{related}: \texttt{"The capital of France is Paris. The capital of Germany is"}
\item \texttt{counterfactual}: \texttt{"The capital of Germany is Munich. The capital of Germany is"}
\item \texttt{short\_irr}: \texttt{"Computers are useful. The capital of Germany is"}
\item \texttt{long\_irr}: \texttt{"$\langle\sim$100-token WikiText-103 paragraph passing the five-level leakage filter$\rangle$ The capital of Germany is"}
\item \texttt{random}: \texttt{"Calculator. The capital of Germany is"}
\end{itemize}

\paragraph{IOI binding/retrieval.}
\begin{itemize}\setlength{\itemsep}{1pt}
\item \texttt{nctx}: \texttt{"When Mary and John went to the store, John gave a drink to"}
\item \texttt{golden}: \texttt{"Mary received the drink. When Mary and John went to the store, John gave a drink to"}
\item \texttt{related}: \texttt{"When Alice and Bob went to the cafe, Bob gave a book to Alice. When Mary and John went to the store, John gave a drink to"}
\item \texttt{counterfactual}: \texttt{"John received the drink. When Mary and John went to the store, John gave a drink to"}
\item \texttt{short\_irr}: \texttt{"Computers are useful. When Mary and John went to the store, John gave a drink to"}
\item \texttt{long\_irr}: \texttt{"$\langle\sim$100-token WikiText-103 paragraph$\rangle$ When Mary and John went to the store, John gave a drink to"}
\item \texttt{random}: \texttt{"Calculator. When Mary and John went to the store, John gave a drink to"}
\end{itemize}
The same paired-comparison invariants hold across both task families: the query, gold token, and model are held fixed within each sample family; only the prepended prefix changes.

\subsection{Length axis}
\label{app:length}

We sweep prefix lengths $\ell\in\{1,12,50,100,300\}$ tokens. The 50-token prefixes truncate long-irrelevant paragraphs at the nearest sentence boundary; 300-token prefixes concatenate two paragraphs under the same leakage filter. Gold-logit shifts and intrusion-head attention to context saturate by $\ell\approx50$; additional content increases diffuse parroting (\S\ref{sec:pathway-parroting}) without further gold suppression.

\subsection{Distractor count axis}
\label{app:count}

To test content dependence against the position-bias alternative in concurrent work~\citep{chowdhury2026circuits}, we repeat the core circuit analyses with distractors appended after the query but before generation, or interleaved between subject and relation. Both preserve the Intrusion-Pathway signature: late-block intrusion-head attention follows the irrelevant content, and gold-logit suppression remains within the prepended default's bootstrap CI. Damage therefore depends on content rather than absolute position.

\subsection{Distractor position axis}
\label{app:position}

We repeat the core circuit analyses with distractors appended after the query, before generation, or interleaved between subject and relation. Late-block intrusion-head attention follows the distractor, and gold-logit suppression remains within the prepended baseline's bootstrap CI.

\subsection{Semantic-distance robustness}
\label{app:semantic_distance}

We test far (physics/mechanics; default \texttt{short\_irr}), medium (geography about another country), and near (same country, different relation) irrelevant prefixes. Intrusion-head attention to context and gold-logit suppression persist across all three distances.

\subsection{Leakage filter}
\label{app:leakage}

For \texttt{long\_irr}, we reject WikiText-103 paragraphs containing whitespace-tokenized substring matches to the sample's gold or subject token, other valid golds or subjects from the same relation, or gold-token semantic neighbors with GloVe-300d cosine similarity $>0.5$~\citep{pennington2014glove}.

\subsection{Distractor pool composition and selection robustness}
\label{app:selection_robust}

The \texttt{short\_irr} pool contains sentences from physics/mechanics, biology, agriculture, everyday objects, weather, and abstract concepts. Sentences have $8$--$18$ whitespace tokens, with no gold-attribute or subject-attribute lexical overlap across the tested LRE relations.

Robustness checks replace per-sample most-damaging selection~\citep{shi2023distracted} with random selection, median-damage selection, or one fixed sentence for all samples. All three preserve the qualitative per-component signature; absolute damage scales monotonically with the expected selection effect.

\FloatBarrier
\section{Intrusion Pathway}
\label{sec:app-pathway}
Conditions retain the main-text notation: \texttt{nctx} (bare query), \texttt{far\_irr} (type-disjoint LRE prefix; neither the query's relation type nor its gold attribute is supplied), and \texttt{long\_irr} (irrelevant WikiText prefix).

\subsection{MLP enrichment}
\label{sec:app-enrichment}
\subsubsection{Subject attributes and preservation}
\label{sec:app-subject}
\begin{figure}[!htbp]
\centering\includegraphics[width=\linewidth]{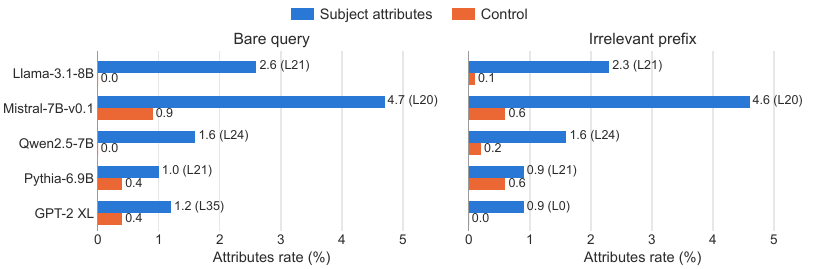}
\caption{\textbf{Subject Attributes Rate.} Percentage of top-20 projected tokens in the Wikipedia bank at the peak layer, with matched controls. Parentheses identify the layer. Values are point estimates.}
\label{fig:app-attributes}
\end{figure}

Subject writes favor their own attribute banks: peak LRE excess is $0.073$--$0.395$ and Wikipedia excess is $0.040$--$0.329$ (Table~\ref{tab:app-enrichment}a). Figure~\ref{fig:app-attributes} compares top-20 Attributes Rates.

\begin{table}[!htbp]
\centering\small
\caption{\textbf{Entity-specific MLP enrichment.} Own-minus-other attribute-bank projection at the last subject or distractor token, read through the unembedding at a frozen final-normalization scale \citep{geva2023dissecting}. Peak selection excludes the final layer; brackets are 95\% paired-bootstrap intervals conditional on the selected layer. Final-layer values are point estimates.}
\label{tab:app-enrichment}
\begin{tabular}{@{}llcrr@{}}
\toprule
Model & Bank & Layer & Peak excess [95\% CI] & Final layer\\
\midrule
\multicolumn{5}{@{}l}{\textit{(a) Subject position}}\\[2pt]
Llama-3.1-8B & LRE & 30 & $+0.151\ [ +0.039,+0.267 ]$ & $+1.217$\\
 & Wikipedia & 30 & $+0.231\ [ +0.178,+0.285 ]$ & $+1.370$\\
Mistral-7B-v0.1 & LRE & 23 & $+0.229\ [ +0.135,+0.318 ]$ & $+0.228$\\
 & Wikipedia & 29 & $+0.186\ [ +0.113,+0.283 ]$ & $+0.488$\\
Qwen2.5-7B & LRE & 24 & $+0.395\ [ +0.289,+0.499 ]$ & $+0.844$\\
 & Wikipedia & 26 & $+0.329\ [ +0.269,+0.394 ]$ & $+1.014$\\
Pythia-6.9B & LRE & 21 & $+0.160\ [ +0.098,+0.226 ]$ & $+0.279$\\
 & Wikipedia & 30 & $+0.198\ [ +0.145,+0.255 ]$ & $+0.467$\\
GPT-2 XL & LRE & 45 & $+0.073\ [ +0.021,+0.129 ]$ & $-0.013$\\
 & Wikipedia & 0 & $+0.040\ [ +0.028,+0.052 ]$ & $-0.044$\\
\midrule
\multicolumn{5}{@{}l}{\textit{(b) Distractor position}}\\[2pt]
Llama-3.1-8B & Category & 30 & $+0.110\ [ +0.075,+0.144 ]$ & $+1.049$\\
 & Wikipedia & 30 & $+0.272\ [ +0.223,+0.321 ]$ & $+1.482$\\
Mistral-7B-v0.1 & Category & 28 & $+0.126\ [ +0.107,+0.145 ]$ & $+0.152$\\
 & Wikipedia & 30 & $+0.165\ [ +0.137,+0.193 ]$ & $+0.555$\\
Qwen2.5-7B & Category & 26 & $+0.386\ [ +0.330,+0.442 ]$ & $+0.922$\\
 & Wikipedia & 26 & $+0.292\ [ +0.239,+0.345 ]$ & $+0.309$\\
Pythia-6.9B & Category & 30 & $+0.173\ [ +0.106,+0.236 ]$ & $+0.353$\\
GPT-2 XL & Category & 38 & $+0.112\ [ +0.088,+0.139 ]$ & $-0.027$\\
 & Wikipedia & 0 & $+0.081\ [ +0.071,+0.093 ]$ & $-0.022$\\
\bottomrule
\end{tabular}
\par\smallskip
\begin{minipage}{\linewidth}\footnotesize
Banks are defined independently of the tested MLP: LRE attributes or Wikipedia content words for subjects; category co-members and category words, or Wikipedia content words, for distractors. Source/context tokens, the gold answer, generic query-category words, and other valid answers are excluded.
\end{minipage}
\end{table}

\subsubsection{Associations at irrelevant-token positions}
\label{sec:app-distractor}
Distractor-position writes also favor their own associations: category excess is $0.110$--$0.386$, with positive Wikipedia comparisons (Table~\ref{tab:app-enrichment}b).

\subsubsection{Contributions of complete MLP writes}
\label{sec:app-mlp-intervention}
\begin{figure}[!htbp]
\centering\includegraphics[width=\linewidth]{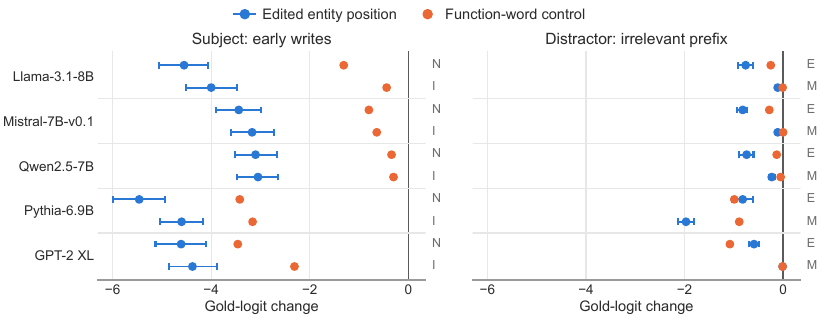}
\caption{\textbf{MLP-write cancellation.} Gold-logit change at END relative to the unedited run of the same prompt. Left: early-band subject edits under bare queries (N) and \texttt{far\_irr} (I). Right: distractor edits in the early (E) or middle (M) band under \texttt{far\_irr}. Bars show 95\% paired-bootstrap intervals for entity-position edits; controls are point estimates at query function words (left) or prefix function words (right).}
\label{fig:app-mlp-intervention}\label{tab:app-mlp-intervention}
\end{figure}

Cancelling early subject writes lowers the gold logit by $3.10$--$5.46$ under bare queries and $3.05$--$4.60$ under \texttt{far\_irr}. Figure~\ref{fig:app-mlp-intervention} includes both subject and distractor edits with function-word controls. These interventions act on complete writes, including content outside the attribute banks.

\subsection{Source-token copying}
\label{sec:app-copying}
\subsubsection{Source substitution with fixed attention}
\label{sec:app-fixed-attention}\label{app:parroting-value-swap}
\begin{figure}[!htbp]
\centering\includegraphics[width=\linewidth]{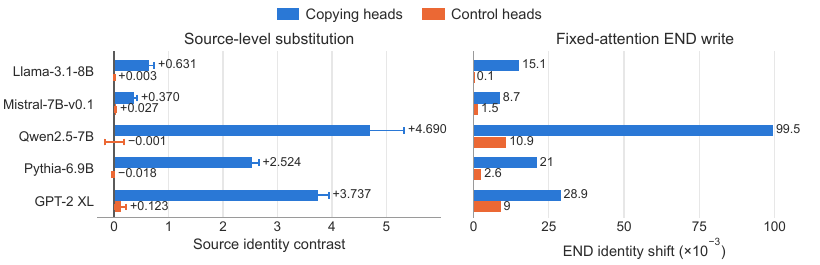}
\caption{\textbf{Identity substitution.} Left: change in the source OV write's replacement-versus-original identity projection, with 95\% confidence intervals. Right: source-value substitution at fixed attention, measured in the END write ($10^{-3}$ units; point estimates). Copying analyses use ten heads per model ranked by own-source OV identity projection minus mean projection onto other sources under \texttt{far\_irr}. The intervention comparison in \S\ref{sec:app-copying-interventions} uses eight heads.}
\label{fig:app-identity}
\end{figure}

Source substitution shifts the source identity contrast by $+0.370$--$+4.690$ across five models and produces a positive END shift when attention is fixed (Figure~\ref{fig:app-identity}). Matched prompts have identical token counts and differ in one token ID, replacing $B$ with $D$ without a matching induction cue.

\subsubsection{Copying after cue removal}
\label{sec:app-cue}\label{app:parroting-cue-removal}
\begin{figure}[!htbp]
\centering\includegraphics[width=\linewidth]{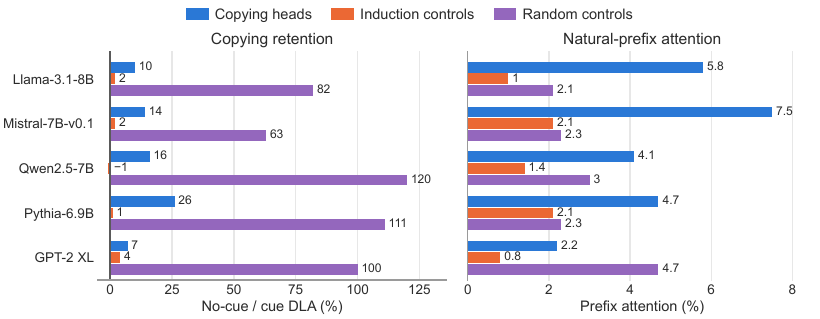}
\caption{\textbf{Cue removal and natural-prefix attention.} Left: median no-cue/cue direct-logit-attribution ratio, keeping $B$, its position, and the final cue fixed. Right: prefix attention in natural prompts without an induction cue. Induction controls are independently selected by prefix matching on repeated random sequences. Random controls are included in both panels. All values are point estimates.}
\label{fig:app-cue}
\end{figure}

Replacing $A\,B\cdots A$ with $C\,B\cdots A$ retains $7$--$26\%$ of copying heads' attribution to $B$; induction controls range from $-1$ to $4\%$ (Figure~\ref{fig:app-cue}). These ratios describe cue sensitivity within the tested heads; they do not compare absolute no-cue copying magnitudes across populations.

\subsubsection{Copying-head interventions across context types}
\label{sec:app-copying-interventions}
The comparison tests eight copying heads under useful and irrelevant prefixes separately and together (Table~\ref{tab:app-copying-interventions}). Under mixed prefixes, edge masking changes the gold logit by $-0.026$ to $-0.145$ across the five models.

\begin{table}[!htbp]
\centering\small
\caption{\textbf{Copying-head interventions.} Gold-logit change relative to the unedited run. Mixed context concatenates the golden and irrelevant prefixes. Whole-head edits zero or donor-replace the head output; END-only zeroing is restricted to END. Prefix-write removal deletes only prefix-source contributions, while prefix-edge masking blocks attention to those sources. Values are point estimates.}
\label{tab:app-copying-interventions}
\setlength{\tabcolsep}{5pt}
\begin{tabular}{@{}llrrrr@{}}
\toprule
Model & Edit & Bare & Golden & Irrelevant & Mixed\\
\midrule
Llama-3.1-8B & Whole-head zeroing & $-0.025$ & $-0.118$ & $-0.034$ & $-0.085$\\
 & Whole-head donor replacement & $-0.003$ & $-0.095$ & $-0.019$ & $-0.062$\\
 & END-only zeroing & $-0.025$ & $-0.095$ & $-0.035$ & $-0.070$\\
 & Prefix-write removal & $0.000$ & $-0.080$ & $-0.009$ & $-0.030$\\
 & Prefix-edge masking & $0.000$ & $-0.068$ & $0.000$ & $-0.035$\\
\addlinespace[2pt]
Mistral-7B-v0.1 & Whole-head zeroing & $-0.016$ & $-0.102$ & $-0.005$ & $-0.084$\\
 & Whole-head donor replacement & $-0.013$ & $-0.100$ & $+0.006$ & $-0.079$\\
 & END-only zeroing & $-0.019$ & $-0.107$ & $-0.010$ & $-0.081$\\
 & Prefix-write removal & $0.000$ & $-0.088$ & $+0.008$ & $-0.060$\\
 & Prefix-edge masking & $0.000$ & $-0.075$ & $+0.010$ & $-0.056$\\
\addlinespace[2pt]
Qwen2.5-7B & Whole-head zeroing & $-0.076$ & $-0.201$ & $-0.085$ & $-0.251$\\
 & Whole-head donor replacement & $-0.036$ & $-0.161$ & $-0.045$ & $-0.168$\\
 & END-only zeroing & $-0.049$ & $-0.180$ & $-0.052$ & $-0.200$\\
 & Prefix-write removal & $0.000$ & $-0.139$ & $-0.021$ & $-0.155$\\
 & Prefix-edge masking & $0.000$ & $-0.133$ & $-0.011$ & $-0.145$\\
\addlinespace[2pt]
Pythia-6.9B & Whole-head zeroing & $+0.027$ & $-0.279$ & $-0.114$ & $-0.232$\\
 & Prefix-write removal & $0.000$ & $-0.015$ & $+0.081$ & $-0.066$\\
 & Prefix-edge masking & $0.000$ & $+0.003$ & $+0.105$ & $-0.072$\\
\addlinespace[2pt]
GPT-2 XL & Whole-head zeroing & $+0.019$ & $+0.034$ & $+0.016$ & $+0.040$\\
 & Prefix-edge masking & $0.000$ & $-0.026$ & $-0.002$ & $-0.026$\\
\bottomrule
\end{tabular}
\end{table}

\subsubsection{Attention-weighted reference for copying}
\label{app:parroting-reference}
For a head $h$, let $\alpha_p=A_h[\mathrm{END},p]$ and let $w_{h,p}=W_O^{(h)}W_V^{(h)}x_p^{(L)}$ denote its unweighted source OV write, omitting normalization and bias terms. Its complete END write and prefix-content contribution are
\begin{equation}
 o_h=\sum_p\alpha_p w_{h,p},\qquad
 o_h^{\mathrm{content}}=\sum_{p\in\mathcal P_{\mathrm{content}}}\alpha_p w_{h,p}.
 \label{eq:parroting-source-writes}
\end{equation}
\label{eq:parroting-content-write}
The content set excludes BOS, stopwords, and query repetitions; attention is not renormalized over it. With source-token unembedding $u_{t_p}$, a shared positive copy-gain reference is
\begin{equation}
\begin{aligned}
 w_{h,p}&=c_hu_{t_p}+\epsilon_{h,p},\qquad c_h>0,\\
 o_h^{\mathrm{content}}&=c_hu_*^{\mathrm{attn}}+r_h,\qquad
 u_*^{\mathrm{attn}}=\sum_{p\in\mathcal P_{\mathrm{content}}}\alpha_pu_{t_p},\quad
 r_h=\sum_{p\in\mathcal P_{\mathrm{content}}}\alpha_p\epsilon_{h,p}.
\end{aligned}
\label{eq:parroting-source-reference}
\end{equation}
\label{eq:parroting-mixture}
The residual includes source-dependent gains and other represented content; it need not be small or orthogonal. A dominant source gives a single-source reference; several contributing sources give a diffuse mixture. For any target direction $u_t$,
\begin{equation}
 \langle o_h^{\mathrm{content}},u_t\rangle
 =c_h\sum_{p\in\mathcal P_{\mathrm{content}}}\alpha_p\langle u_{t_p},u_t\rangle+\langle r_h,u_t\rangle.
 \label{eq:parroting-target-projection}
\end{equation}
Thus contributions depend on token geometry as well as attention. The held-out comparison permutes identity--attention pairings while preserving both multisets:
\begin{equation}
 u_*^{\mathrm{shuffle}}=\sum_{p\in\mathcal P_{\mathrm{content}}}\alpha_pu_{t_{\pi(p)}}.
 \label{eq:parroting-shuffled-reference}
\end{equation}
Equal weights can make the shuffle uninformative. This tests an identity component of the write. CSPA \citep{mcdougall2023copy} instead preserves negative source-wise projections with destination-prediction filtering, evaluated against mean ablation.

\subsection{Multi-source copying and association output}
\label{sec:app-transport}
\subsubsection{Attention-weighted source mixtures}
\label{sec:app-multisource}\label{app:diffuse-reconstruction}
Correct identity--attention pairing improves held-out reconstruction for $2$--$7$ of ten selected copying heads per model, compared with $0$--$1$ control heads (Table~\ref{tab:app-multisource}). The reported cosine ranges describe alignment with this component.

\begin{table}[!htbp]
\centering\small
\caption{\textbf{Multi-source reconstruction.} Number of heads whose paired-bootstrap interval for shuffled-minus-correct reconstruction error is above zero. Cosines give alignment between copying writes and the identity mixture.}
\label{tab:app-multisource}
\begin{tabular}{@{}lccc@{}}
\toprule
Model & Copying heads & Control heads & Cosine range\\
\midrule
Llama-3.1-8B & 2/10 & 1/10 & 0.03--0.23\\
Mistral-7B-v0.1 & 3/10 & 1/10 & 0.06--0.18\\
Qwen2.5-7B & 3/10 & 0/10 & 0.06--0.25\\
Pythia-6.9B & 6/10 & 0/10 & 0.03--0.15\\
GPT-2 XL & 7/10 & 1/10 & 0.07--0.20\\
\bottomrule
\end{tabular}
\end{table}

\subsubsection{Association-related output}
\label{sec:app-association}
Copying writes show positive association excess in all five models; held-out profile errors and cosines measure source-profile preservation separately (Table~\ref{tab:app-association}).

\begin{table}[!htbp]
\centering\small
\caption{\textbf{Association output through OV.} Copying-minus-identity-reference association projection (excluding the source token), with 95\% intervals and layer/head-matched random controls. Profile error and cosine compare source/output association coordinates; one scalar gain is fitted on the discovery half and fixed on held-out data. Profiles are held-out medians; control excesses are point estimates.}
\label{tab:app-association}
\begin{tabular}{@{}lrrcc@{}}
\toprule
 & \multicolumn{2}{c}{Association excess} & \multicolumn{2}{c}{Profile alignment}\\
\cmidrule(lr){2-3}\cmidrule(l){4-5}
Model & Copying heads [95\% CI] & Control & Rel.\ error & Cosine\\
\midrule
Llama-3.1-8B & $+0.017\ [+0.009,+0.025]$ & $+0.001$ & 1.000 & $+0.005$\\
Mistral-7B-v0.1 & $+0.008\ [+0.005,+0.011]$ & $+0.001$ & 1.025 & $+0.024$\\
Qwen2.5-7B & $+0.086\ [+0.048,+0.127]$ & $-0.005$ & 0.989 & $+0.067$\\
Pythia-6.9B & $+0.095\ [+0.079,+0.112]$ & $+0.003$ & 0.996 & $+0.087$\\
GPT-2 XL & $+0.161\ [+0.133,+0.190]$ & $-0.028$ & 1.000 & $-0.011$\\
\bottomrule
\end{tabular}
\end{table}

\subsubsection{Selective edits to association content}
\label{sec:app-association-edit}
Table~\ref{tab:app-association-edit} removes a rank-12 association component with the identity direction projected out and attention fixed. Identity-removal and matched random edits are controls; identity and final-answer readouts accompany the association changes.

\begin{table}[!htbp]
\centering\small
\caption{\textbf{Association-subspace removal.} (a) Association-output change; brackets are 95\% intervals, rounded to four decimals. Random subspaces match rank and norm. (b) Identity projection and final gold logit before/after association removal. Controls and before/after values are point estimates.}
\label{tab:app-association-edit}
\begin{tabular}{@{}lrrr@{}}
\toprule
\multicolumn{4}{@{}l}{\textit{(a) Association-output change}}\\[3pt]
Model & Association removal [95\% CI] & Identity removal & Random subspace\\
\midrule
Llama-3.1-8B & $-0.0005\ [-0.0009,-0.0000]$ & $+0.0010$ & $-0.0001$\\
Mistral-7B-v0.1 & $+0.0001\ [+0.0001,+0.0001]$ & $+0.0000$ & $-0.0000$\\
Qwen2.5-7B & $-0.0035\ [-0.0052,-0.0019]$ & $+0.0006$ & $+0.0002$\\
Pythia-6.9B & $-0.0055\ [-0.0078,-0.0032]$ & $+0.0000$ & $-0.0004$\\
GPT-2 XL & $-0.0163\ [-0.0214,-0.0116]$ & $-0.0041$ & $+0.0001$\\
\bottomrule
\end{tabular}
\par\smallskip
\begin{tabular}{@{}lrrrr@{}}
\toprule
\multicolumn{5}{@{}l}{\textit{(b) Identity and answer readouts}}\\[3pt]
 & \multicolumn{2}{c}{Identity projection} & \multicolumn{2}{c}{Gold logit}\\
\cmidrule(lr){2-3}\cmidrule(l){4-5}
Model & Before & After & Before & After\\
\midrule
Llama-3.1-8B & 0.387 & 0.384 & 18.921 & 18.909\\
Mistral-7B-v0.1 & 0.206 & 0.206 & 15.188 & 15.184\\
Qwen2.5-7B & 3.041 & 3.033 & 18.529 & 18.519\\
Pythia-6.9B & 1.573 & 1.552 & 16.209 & 16.233\\
GPT-2 XL & 2.113 & 2.085 & 12.816 & 12.807\\
\bottomrule
\end{tabular}
\end{table}

\subsection{Relational readers: frame and answer type}
\label{sec:app-readers}
\begin{figure}[!htbp]
\centering\includegraphics[width=\linewidth]{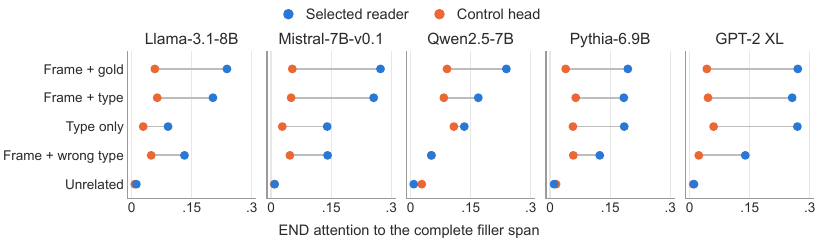}
\caption{\textbf{Reader attention by frame and type.} END attention summed over the complete filler span at the end of the prefix. ``Type'' denotes an incorrect answer of the right type. Readers are the ten middle-layer heads selected on the discovery half by golden-answer-span attention minus maximum \texttt{far\_irr} context attention. All values, including control heads, are point estimates.}
\label{fig:app-readers}\label{tab:app-readers}
\end{figure}

Unrelated-filler attention is $0.008$--$0.012$, compared with $0.194$--$0.273$ for framed gold answers (Figure~\ref{fig:app-readers}). The controlled alternatives separate answer type from relational frame.

\subsection{Restoring attention writes and residual states}
\label{sec:app-restoration}
\subsubsection{Middle-block attention and MLP outputs}
\label{sec:app-sublayer-restoration}
We replace END outputs in the middle depth third (approximately $33$--$67\%$) with matched bare-query outputs, retaining the prefixed input. For gold logit or gold-versus-best-competitor margin $s$, the signed restoration ratio is
\begin{equation}
 R_s=100\,\frac{s_{\mathrm{restored}}-s_{\mathrm{ctx}}}{s_{\mathrm{nctx}}-s_{\mathrm{ctx}}}.
 \label{eq:app-recovery}
\end{equation}
When context lowers $s$, $100\%$ denotes full recovery. When context raises $s$, a positive ratio denotes removal of that benefit. Table~\ref{tab:app-restoration} retains the baseline changes so both cases are visible; Qwen's \texttt{far\_irr} gold and margin baselines are positive. Golden-context results use the same definition (Table~\ref{tab:app-golden}).

\begin{table}[!htbp]
\centering\small
\caption{\textbf{Middle-block attention restoration.} Baseline changes are context minus bare-query values. Signed ratios use Eq.~\eqref{eq:app-recovery}; brackets are 95\% confidence intervals. For Qwen \texttt{far\_irr}, the positive ratio removes a baseline gain.}
\label{tab:app-restoration}
\begin{tabular}{@{}llrrrr@{}}
\toprule
 & & \multicolumn{2}{c}{Baseline change} & \multicolumn{2}{c}{Signed ratio (\%) [95\% CI]}\\
\cmidrule(lr){3-4}\cmidrule(l){5-6}
Model & Condition & Gold logit & Margin & Gold logit & Margin\\
\midrule
Llama-3.1-8B & \texttt{far\_irr} & $-1.450$ & $-0.275$ & 61 [57,65] & 87 [67,120]\\
 & \texttt{long\_irr} & $-1.882$ & $-1.196$ & 54 [50,58] & 70 [64,78]\\
Mistral-7B-v0.1 & \texttt{far\_irr} & $-0.708$ & $-0.107$ & 57 [52,62] & 168 [$-388$,988]\\
 & \texttt{long\_irr} & $-1.565$ & $-0.731$ & 60 [56,64] & 79 [69,92]\\
Qwen2.5-7B & \texttt{far\_irr} & $+0.515$ & $+0.330$ & 72 [56,96] & 57 [36,110]\\
 & \texttt{long\_irr} & $-0.477$ & $-0.788$ & $-6$ [$-56$,17] & 17 [1,32]\\
Pythia-6.9B & \texttt{far\_irr} & $-0.525$ & $-0.528$ & 90 [77,109] & 103 [87,130]\\
 & \texttt{long\_irr} & $-1.227$ & $-0.897$ & 71 [65,78] & 86 [74,100]\\
GPT-2 XL & \texttt{far\_irr} & $-0.376$ & $-0.324$ & 41 [21,53] & 72 [54,93]\\
 & \texttt{long\_irr} & $-1.724$ & $-1.263$ & 53 [48,58] & 66 [58,75]\\
\bottomrule
\end{tabular}
\end{table}

\begin{figure}[!htbp]
\centering\includegraphics[width=\linewidth]{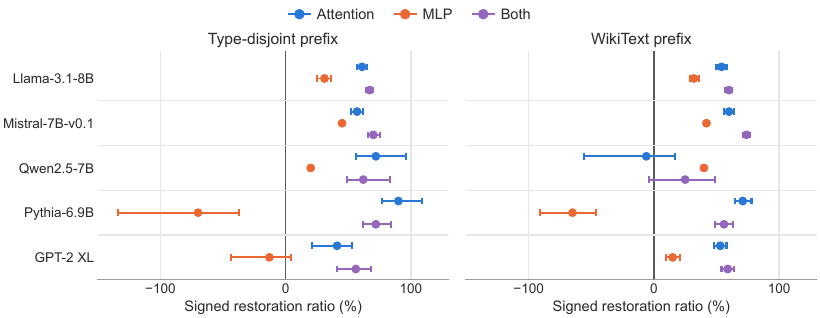}
\caption{\textbf{Sublayer restoration in the middle third.} Signed gold-logit restoration ratio. Bars are 95\% intervals. Each point is a separate intervention, so ratios are not additive. Qwen \texttt{far\_irr} ratios measure removal of a context benefit.}
\label{fig:app-sublayers}
\end{figure}

Figure~\ref{fig:app-sublayers} compares attention, MLP, and joint restoration in the same middle window. The modelwise panels retain each signed effect and its reported uncertainty.

\begin{table}[!htbp]
\centering\small
\caption{\textbf{The same restoration under golden context.} Benefit is the unedited gold-logit increase over the bare query; the removed fraction includes 95\% confidence intervals.}
\label{tab:app-golden}
\begin{tabular}{@{}lrr@{}}
\toprule
Model & Gold-logit benefit & Benefit removed (\%) [95\% CI]\\
\midrule
Llama-3.1-8B & $+0.767$ & 68 [61,79]\\
Mistral-7B-v0.1 & $+0.502$ & 58 [48,69]\\
Qwen2.5-7B & $+0.955$ & 33 [23,43]\\
Pythia-6.9B & $+0.690$ & 45 [30,57]\\
GPT-2 XL & $+0.683$ & 87 [70,109]\\
\bottomrule
\end{tabular}
\end{table}

\subsubsection{Restoration across depth}
\label{sec:app-depth-restoration}
\begin{figure}[!htbp]
\centering\includegraphics[width=\linewidth]{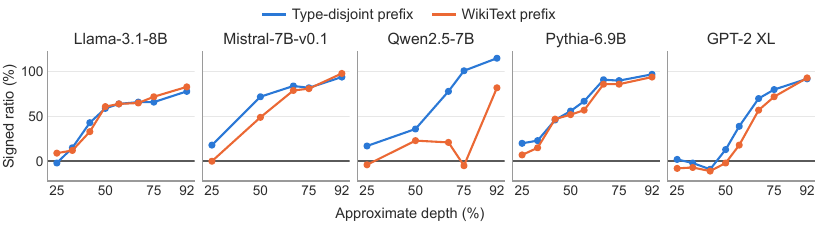}
\caption{\textbf{Single-layer END-residual restoration.} Signed gold-logit restoration ratio. Points are the reported layers nearest the indicated depth fractions; segments connect observations without smoothing. The denominator convention is Eq.~\eqref{eq:app-recovery}.}
\label{fig:app-depth}
\end{figure}

Table~\ref{tab:app-windows} compares complete early, middle, and late depth thirds. Figure~\ref{fig:app-depth} replaces the END residual at individual layers, with subsequent layers continuing on the prefixed run. The two experiments vary the restored site and intervention scope separately.

\begin{table}[!htbp]
\centering\small
\caption{\textbf{Window restoration.} Signed gold-logit restoration ratio (\%) after replacing both attention and MLP outputs at END within the early, middle, or late depth third. Brackets give 95\% confidence intervals. Windows are restored in separate runs; their ratios do not partition the total effect. Qwen \texttt{far\_irr} ratios measure removal of a context benefit.}
\label{tab:app-windows}
\begin{tabular}{@{}llrrr@{}}
\toprule
Model & Condition & Early & Middle & Late\\
\midrule
Llama-3.1-8B & \texttt{far\_irr} & 6 [2,9] & 67 [64,70] & 94 [91,97]\\
 & \texttt{long\_irr} & 1 [$-2$,4] & 60 [57,62] & 88 [85,90]\\
Mistral-7B-v0.1 & \texttt{far\_irr} & 2 [$-5$,8] & 70 [66,75] & 98 [93,103]\\
 & \texttt{long\_irr} & 4 [$-1$,8] & 74 [71,77] & 90 [88,93]\\
Qwen2.5-7B & \texttt{far\_irr} & 17 [8,27] & 62 [49,83] & 95 [90,99]\\
 & \texttt{long\_irr} & 22 [7,47] & 25 [$-4$,49] & 107 [102,116]\\
Pythia-6.9B & \texttt{far\_irr} & 12 [$-2$,24] & 72 [62,84] & 115 [96,141]\\
 & \texttt{long\_irr} & 1 [$-9$,8] & 56 [49,63] & 87 [78,98]\\
GPT-2 XL & \texttt{far\_irr} & $-3$ [$-16$,7] & 56 [41,68] & 81 [61,107]\\
 & \texttt{long\_irr} & 0 [$-7$,5] & 59 [54,64] & 69 [62,76]\\
\bottomrule
\end{tabular}
\end{table}
\section{Intrusion heads}
\label{sec:app-intrusion-heads}
\providecommand{\appest}[2]{\shortstack{$#1$\\[-1pt]{\scriptsize $#2$}}}
\subsection{Populations and contextual writes}
\label{sec:app-intrusion}
\subsubsection{Populations and attention}
\label{sec:app-intrusion-populations}
We freeze three distinct head populations: $I$ contains the twenty highest intrusion-scored heads; $I^\star$ contains twenty heads selected after excluding the top forty retrieval-scored heads; $I_{\rm only}=I_{20}\setminus R_{20}$ is used only in \S\ref{sec:app-contextual-retrieval}. These sets are distinct from downstream $K{=}10$ interventions. Short and long denote \texttt{far\_irr} and \texttt{long\_irr}.

\begin{table}[!htbp]
\setlength{\tabcolsep}{4pt}
\centering\footnotesize
\caption{\textbf{Population stability and attention shifts.} Depth counts and stability describe $I$; inclusion and Jaccard use relation-stratified discovery resamples. Attention shifts describe $I^\star$ under short prefixes relative to bare queries and show selected source regions.}
\label{tab:app-intrusion-populations}
\begin{tabular}{@{}lcccrrr@{}}
\toprule
Model & \shortstack{Early/mid/late\\($I$)} & \shortstack{Inclusion\\$\geq0.8$} & Jaccard & $\Delta a_{\rm prefix}$ & $\Delta a_{\rm BOS}$ & $\Delta a_{\rm rel}$\\
\midrule
Llama-3.1-8B & 1/2/17 & 13/20 & 0.666 & $+0.09$ & $-0.08$ & $-0.01$\\
Mistral-7B-v0.1 & 0/4/16 & 12/20 & 0.662 & $+0.10$ & $-0.10$ & $-0.00$\\
Qwen2.5-7B & 0/1/19 & 14/20 & 0.615 & $+0.23$ & $-0.09$ & $-0.09$\\
Pythia-6.9B & 0/6/14 & 10/20 & 0.487 & $+0.14$ & $-0.09$ & $-0.03$\\
GPT-2 XL & 2/8/10 & 12/20 & 0.587 & $+0.25$ & $-0.11$ & $-0.09$\\
\bottomrule
\end{tabular}
\end{table}

\subsubsection{Gold and distractor support}
\label{sec:app-intrusion-writing}
\begin{figure}[!htbp]\centering\includegraphics[width=\linewidth]{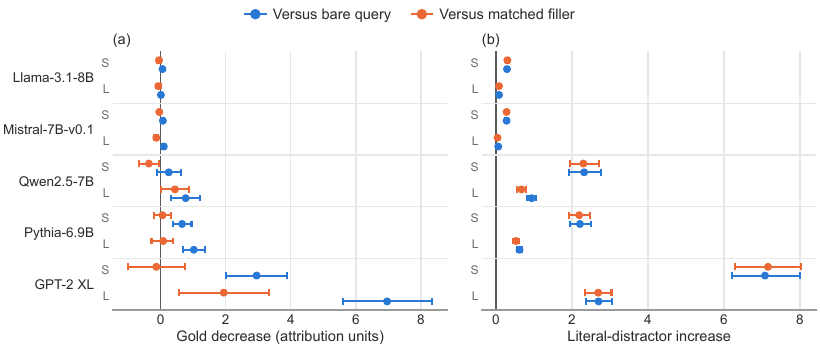}
\caption{\textbf{Reference-controlled attribution in $I^\star$.} Each row shows short (S) or long (L) prefixes. Gold decrease is reference minus irrelevant; distractor increase uses the reverse contrast. Readouts sum heads with fixed final-normalization weights. Bars are 95\% intervals; axes are shared across models within each readout.}\label{tab:app-intrusion-contrasts}\label{fig:app-intrusion-contrasts}
\end{figure}

Literal-distractor attribution rises against both bare and token-exact neutral-filler references (Figure~\ref{tab:app-intrusion-contrasts}). The gold contrast depends on the reference: the filler comparison retains a positive decrease for the long-prefix Qwen and GPT-2 XL conditions. Table~\ref{tab:app-intrusion-vocabulary} gives the separate vocabulary-controlled readouts for $I$.

\begin{table}[!htbp]
\setlength{\tabcolsep}{4pt}
\centering\footnotesize
\caption{\textbf{Vocabulary-controlled attribution in $I$.} Irrelevant minus filler, after subtracting frequency- and norm-matched vocabulary controls. These point estimates use the unfiltered population; they are separate from the $I^\star$ contrasts above.}
\label{tab:app-intrusion-vocabulary}
\begin{tabular}{@{}lrrrrr@{}}
\toprule
Model & Literal tokens & Distractor category & Gold & Query category & Other answers\\
\midrule
Llama-3.1-8B & $+0.2794$ & $+0.1182$ & $+0.0515$ & $+0.0277$ & $+0.0453$\\
Mistral-7B-v0.1 & $+0.2881$ & $+0.1080$ & $+0.2178$ & $+0.0158$ & $+0.0138$\\
Qwen2.5-7B & $+0.1440$ & $+0.0233$ & $+0.0101$ & $-0.0153$ & $-0.0062$\\
Pythia-6.9B & $+0.2778$ & $+0.0970$ & $+0.0792$ & $+0.0230$ & $+0.0202$\\
GPT-2 XL & $+0.2428$ & $+0.1349$ & $-0.0092$ & $+0.0121$ & $+0.0089$\\
\bottomrule
\end{tabular}
\end{table}

\subsection{Intrusion writes and interventions}
\label{sec:app-intrusion-interventions}
\subsubsection{Prefix and non-prefix contributions}
\label{sec:app-intrusion-sources}
The split $o_h=P_h+U_h$ separates prefix and remaining-source contributions. Prefix writes account for about 52\%, 32\%, 94\%, 68\%, and 74\% of increased literal-distractor attribution, respectively. The net gold reduction lies in non-prefix terms; the END-position term is largest in Llama, Mistral, and Qwen (Table~\ref{tab:app-intrusion-sources}).

\begin{table}[!htbp]
\setlength{\tabcolsep}{4pt}
\centering\footnotesize
\caption{\textbf{Source-resolved attribution in $I^\star$.} Short irrelevant prefix minus bare query, with means of head-summed readouts. $P$ is the prefix contribution; $\Delta U$ partitions changes at the remaining sources. Entries are rounded independently.}
\label{tab:app-intrusion-sources}
\begin{tabular}{@{}llrrrrrr@{}}
\toprule
Model & Target & Total & $P$ & $\Delta U_{\rm subj}$ & $\Delta U_{\rm rel}$ & $\Delta U_{\rm END}$ & $\Delta U_{\rm BOS}$\\
\midrule
Llama-3.1-8B & Gold & $-0.06$ & $-0.00$ & $+0.04$ & $-0.03$ & $-0.07$ & $-0.00$\\
 & $D_{\rm lit}$ & $+0.29$ & $+0.15$ & $+0.03$ & $+0.10$ & $+0.02$ & $+0.00$\\
Mistral-7B-v0.1 & Gold & $-0.07$ & $-0.01$ & $+0.02$ & $+0.05$ & $-0.12$ & $+0.00$\\
 & $D_{\rm lit}$ & $+0.28$ & $+0.09$ & $+0.03$ & $+0.15$ & $+0.01$ & $+0.00$\\
Qwen2.5-7B & Gold & $-0.25$ & $+0.07$ & $-0.06$ & $+0.05$ & $-0.33$ & $+0.02$\\
 & $D_{\rm lit}$ & $+2.32$ & $+2.19$ & $+0.04$ & $+0.09$ & $+0.01$ & $-0.01$\\
Pythia-6.9B & Gold & $-0.66$ & $+0.00$ & $-0.28$ & $-0.23$ & $-0.15$ & $-0.01$\\
 & $D_{\rm lit}$ & $+2.21$ & $+1.51$ & $+0.17$ & $+0.47$ & $+0.07$ & $+0.00$\\
GPT-2 XL & Gold & $-2.95$ & $-0.04$ & $-1.08$ & $-1.41$ & $-0.37$ & $-0.06$\\
 & $D_{\rm lit}$ & $+7.08$ & $+5.22$ & $+0.48$ & $+1.30$ & $+0.16$ & $-0.08$\\
\bottomrule
\end{tabular}
\end{table}

\subsubsection{Clean-write restoration}
\label{sec:app-intrusion-restoration}
\begin{figure}[!htbp]\centering\includegraphics[width=\linewidth]{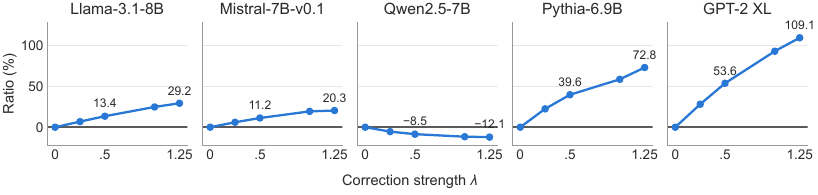}
\caption{\textbf{Correction-strength sweep.} Point estimates for $o_h^{\rm ctx}+\lambda(o_h^{\rm nctx}-o_h^{\rm ctx})$. All five panels use the same ratio axis. The sweep is separate from Table~\ref{tab:app-intrusion-restoration}.}\label{fig:app-intrusion-dose}
\end{figure}

Table~\ref{tab:app-intrusion-restoration} replaces $I$'s END writes with bare-query donors while retaining the prefix. It reports normalized movement toward the no-context gold logit. Figure~\ref{fig:app-intrusion-dose} varies replacement strength in a separately reported sweep.

\begin{table}[!htbp]
\setlength{\tabcolsep}{4pt}
\centering\footnotesize
\caption{\textbf{No-context write restoration and controls.} Point estimates. The ratio is $100(\ell_{\rm edited}-\ell_{\rm ctx})/(\ell_{\rm nctx}-\ell_{\rm ctx})$. Controls contain twenty random, layer-, score-decile-, or prefix-attention-decile-matched heads. A positive value denotes movement toward the bare-query logit, including removal of helpful-context gain when context raises that logit.}
\label{tab:app-intrusion-restoration}
\begin{tabular}{@{}llrrrrr@{}}
\toprule
Model & Context & $I$ (20) & Random & Layer & Score & Attention\\
\midrule
Llama-3.1-8B & Short irrelevant & $24.2$ & $1.9$ & $6.6$ & $7.3$ & $2.2$\\
 & Long irrelevant & $22.5$ & $-0.1$ & $7.2$ & $6.0$ & $1.0$\\
 & Golden & $15.6$ & $-3.0$ & $-1.3$ & $6.2$ & $4.2$\\
\addlinespace
Mistral-7B-v0.1 & Short irrelevant & $17.9$ & $3.6$ & $7.1$ & $10.8$ & $5.8$\\
 & Long irrelevant & $22.6$ & $3.7$ & $6.1$ & $8.0$ & $3.5$\\
 & Golden & $1.7$ & $-0.2$ & $8.1$ & $16.4$ & $1.2$\\
\addlinespace
Qwen2.5-7B & Short irrelevant & $-12.0$ & $-0.5$ & $-2.1$ & $6.2$ & $9.5$\\
 & Long irrelevant & $73.6$ & $-5.8$ & $5.9$ & $12.5$ & $1.0$\\
 & Golden & $41.3$ & $0.8$ & $10.3$ & $26.3$ & $9.3$\\
\addlinespace
Pythia-6.9B & Short irrelevant & $57.8$ & $2.9$ & $9.2$ & $23.0$ & $11.9$\\
 & Long irrelevant & $31.7$ & $2.1$ & $4.7$ & $14.3$ & $2.9$\\
 & Golden & $15.7$ & $-5.6$ & $-1.9$ & $15.6$ & $-0.5$\\
\addlinespace
GPT-2 XL & Short irrelevant & $93.0$ & $4.7$ & $16.0$ & $40.4$ & $6.5$\\
 & Long irrelevant & $32.1$ & $2.2$ & $6.0$ & $13.5$ & $1.5$\\
 & Golden & $14.0$ & $-1.0$ & $3.6$ & $16.6$ & $5.5$\\
\bottomrule
\end{tabular}
\end{table}

\subsubsection{Prefix removal and non-prefix amplification}
\label{sec:app-four-writes}
\begin{figure}[!htbp]\centering\includegraphics[width=\linewidth]{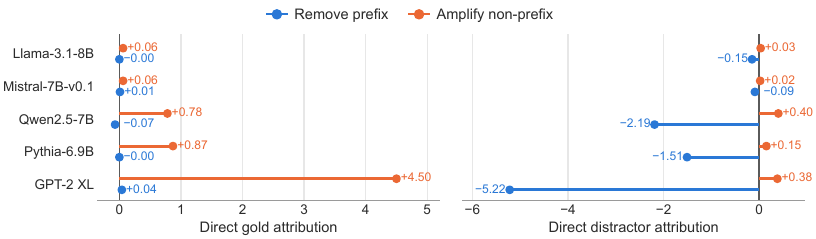}
\caption{\textbf{Immediate effects on edited writes.} Short-prefix direct-attribution changes summed over $I^\star$, averaged over held-out facts. Points show the individual removal and amplification terms; they are distinct from final prediction changes.}\label{fig:app-write-components}
\end{figure}

For fixed head inputs and prefix attention mass $a_h<1$, the four replacements are
\begin{equation}
\begin{aligned}
 o_h^{\rm remove}&=U_h,&o_h^{\rm amplify}&=P_h+\frac{U_h}{1-a_h},\\
 o_h^{\rm both}&=\frac{U_h}{1-a_h},&o_h^{\rm gain}&=\frac{P_h+U_h}{1-a_h}.
\end{aligned}\label{eq:app-four-writes}
\end{equation}
The first two terms separate the local effects of prefix removal and attention renormalization. Figure~\ref{fig:app-write-components} measures their immediate attributions; Table~\ref{tab:app-four-writes} measures the resulting final predictions. In Mistral, whole-write gain increases gold log-probability more than the combined edit; its short-prefix distractor logit also increases.

\begin{table}[!htbp]
\setlength{\tabcolsep}{4pt}
\centering\footnotesize
\caption{\textbf{Four-write interventions in $I^\star$.} Each cell gives gold log-probability change / strongest literal-distractor logit change, with the gold 95\% interval beneath. Changes are relative to the unedited run. Each column is a separate intervention; effects are not additive.}
\label{tab:app-four-writes}
\begin{tabular}{@{}llcccc@{}}
\toprule
Model & Prefix & \shortstack{Remove\\prefix} & \shortstack{Amplify\\non-prefix} & Both & \shortstack{Whole-write\\gain}\\
\midrule
Llama-3.1-8B & Short & \shortstack{$+0.004$\,/\,$-0.14$\\[-1pt]{\scriptsize$[-0.004, +0.014]$}} & \shortstack{$+0.002$\,/\,$-0.01$\\[-1pt]{\scriptsize$[-0.001, +0.005]$}} & \shortstack{$+0.006$\,/\,$-0.14$\\[-1pt]{\scriptsize$[-0.001, +0.014]$}} & \shortstack{$+0.002$\,/\,$+0.02$\\[-1pt]{\scriptsize$[-0.004, +0.007]$}}\\
\addlinespace[1pt]
 & Long & \shortstack{$-0.018$\,/\,$-0.01$\\[-1pt]{\scriptsize$[-0.031, -0.005]$}} & \shortstack{$-0.002$\,/\,$-0.00$\\[-1pt]{\scriptsize$[-0.019, +0.009]$}} & \shortstack{$-0.022$\,/\,$-0.01$\\[-1pt]{\scriptsize$[-0.041, -0.005]$}} & \shortstack{$-0.033$\,/\,$-0.34$\\[-1pt]{\scriptsize$[-0.068, +0.001]$}}\\
\addlinespace[1pt]

Mistral-7B-v0.1 & Short & \shortstack{$-0.003$\,/\,$-0.22$\\[-1pt]{\scriptsize$[-0.007, -0.000]$}} & \shortstack{$+0.014$\,/\,$+0.00$\\[-1pt]{\scriptsize$[+0.010, +0.019]$}} & \shortstack{$+0.011$\,/\,$-0.21$\\[-1pt]{\scriptsize$[+0.007, +0.016]$}} & \shortstack{$+0.017$\,/\,$+0.06$\\[-1pt]{\scriptsize$[+0.012, +0.022]$}}\\
\addlinespace[1pt]
 & Long & \shortstack{$-0.016$\,/\,$-0.06$\\[-1pt]{\scriptsize$[-0.024, -0.008]$}} & \shortstack{$+0.029$\,/\,$-0.01$\\[-1pt]{\scriptsize$[+0.022, +0.037]$}} & \shortstack{$+0.017$\,/\,$-0.07$\\[-1pt]{\scriptsize$[+0.010, +0.025]$}} & \shortstack{$+0.038$\,/\,$-0.15$\\[-1pt]{\scriptsize$[+0.020, +0.056]$}}\\
\addlinespace[1pt]

Qwen2.5-7B & Short & \shortstack{$-0.013$\,/\,$-0.49$\\[-1pt]{\scriptsize$[-0.022, -0.003]$}} & \shortstack{$+0.012$\,/\,$+0.03$\\[-1pt]{\scriptsize$[+0.005, +0.020]$}} & \shortstack{$+0.002$\,/\,$-0.45$\\[-1pt]{\scriptsize$[-0.008, +0.014]$}} & \shortstack{$-0.086$\,/\,$+0.50$\\[-1pt]{\scriptsize$[-0.264, +0.015]$}}\\
\addlinespace[1pt]
 & Long & \shortstack{$-0.064$\,/\,$-0.36$\\[-1pt]{\scriptsize$[-0.090, -0.036]$}} & \shortstack{$+0.035$\,/\,$-0.07$\\[-1pt]{\scriptsize$[+0.015, +0.057]$}} & \shortstack{$-0.016$\,/\,$-0.41$\\[-1pt]{\scriptsize$[-0.049, +0.016]$}} & \shortstack{$-0.831$\,/\,$+0.05$\\[-1pt]{\scriptsize$[-1.404, -0.391]$}}\\
\addlinespace[1pt]

Pythia-6.9B & Short & \shortstack{$+0.026$\,/\,$-0.34$\\[-1pt]{\scriptsize$[+0.017, +0.035]$}} & \shortstack{$+0.008$\,/\,$-0.02$\\[-1pt]{\scriptsize$[+0.002, +0.014]$}} & \shortstack{$+0.032$\,/\,$-0.35$\\[-1pt]{\scriptsize$[+0.021, +0.043]$}} & \shortstack{$-2.27$\,/\,$+2.03$\\[-1pt]{\scriptsize$[-4.44, -0.59]$}}\\
\addlinespace[1pt]
 & Long & \shortstack{$+0.051$\,/\,$-0.12$\\[-1pt]{\scriptsize$[+0.031, +0.076]$}} & \shortstack{$+0.013$\,/\,$-0.03$\\[-1pt]{\scriptsize$[-0.001, +0.030]$}} & \shortstack{$+0.064$\,/\,$-0.14$\\[-1pt]{\scriptsize$[+0.034, +0.097]$}} & \shortstack{$-5.10$\,/\,$+1.51$\\[-1pt]{\scriptsize$[-7.43, -3.03]$}}\\
\addlinespace[1pt]

GPT-2 XL & Short & \shortstack{$+0.044$\,/\,$-0.57$\\[-1pt]{\scriptsize$[+0.016, +0.074]$}} & \shortstack{$+0.043$\,/\,$-0.08$\\[-1pt]{\scriptsize$[+0.027, +0.064]$}} & \shortstack{$+0.080$\,/\,$-0.62$\\[-1pt]{\scriptsize$[+0.043, +0.122]$}} & \shortstack{$+0.017$\,/\,$+0.60$\\[-1pt]{\scriptsize$[-0.020, +0.064]$}}\\
\addlinespace[1pt]
 & Long & \shortstack{$+0.156$\,/\,$-0.30$\\[-1pt]{\scriptsize$[+0.104, +0.213]$}} & \shortstack{$+0.139$\,/\,$-0.05$\\[-1pt]{\scriptsize$[+0.100, +0.182]$}} & \shortstack{$+0.277$\,/\,$-0.34$\\[-1pt]{\scriptsize$[+0.197, +0.371]$}} & \shortstack{$-0.281$\,/\,$+0.48$\\[-1pt]{\scriptsize$[-0.414, -0.157]$}}\\
\addlinespace[1pt]
\bottomrule
\end{tabular}
\end{table}

\subsubsection{Gold and distractor effects of the same intervention}
\label{sec:app-two-sided-mask}
\begin{figure}[!htbp]\centering\includegraphics[width=\linewidth]{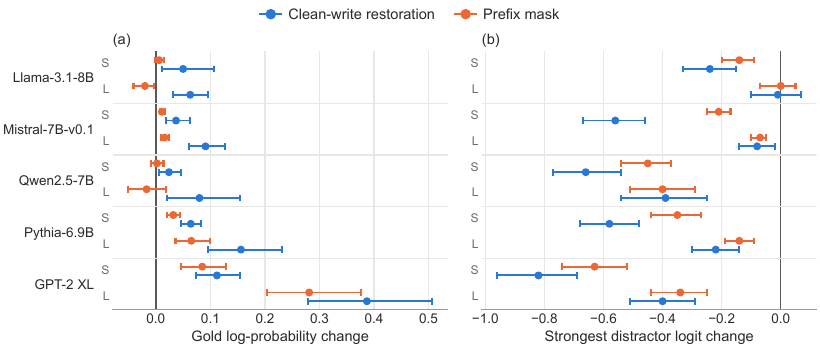}
\caption{\textbf{Same-pass outcomes in $I^\star$.} Restore uses bare-query END-write donors; mask blocks END-to-prefix attention without a donor. Changes are relative to the unedited irrelevant-context run, with 95\% intervals. Short (S) and long (L) prefixes share axes. Gold log-probability and distractor logits are separate metrics.}\label{tab:app-two-sided-mask}\label{fig:app-two-sided-mask}
\end{figure}

Figure~\ref{tab:app-two-sided-mask} pairs the answer and distractor readouts from each forward pass. Masking increases gold log-probability in both prefix conditions for Mistral, Pythia, and GPT-2 XL; Llama's long-prefix change is negative and Qwen's intervals include zero. Table~\ref{tab:app-mask-controls} evaluates golden context and matched heads.

\begin{table}[!htbp]
\setlength{\tabcolsep}{4pt}\centering\footnotesize
\caption{\textbf{Joint improvement by fact.} Percentages with increased gold log-probability and decreased distractor probability / strongest distractor logit. Entries use the same passes as Figure~\ref{tab:app-two-sided-mask}.}\label{tab:app-joint-success}
\begin{tabular}{@{}lrrrr@{}}\toprule
&\multicolumn{2}{c}{Short}&\multicolumn{2}{c}{Long}\\
Model&Restore&Mask&Restore&Mask\\\midrule
Llama-3.1-8B&72 / 59&46 / 47&67 / 37&41 / 30\\
Mistral-7B-v0.1&79 / 84&81 / 52&83 / 48&73 / 32\\
Qwen2.5-7B&50 / 44&44 / 28&44 / 34&28 / 14\\
Pythia-6.9B&79 / 77&74 / 74&67 / 63&70 / 60\\
GPT-2 XL&88 / 84&84 / 84&77 / 71&80 / 71\\\bottomrule
\end{tabular}\end{table}

\begin{table}[!htbp]
\setlength{\tabcolsep}{4pt}
\centering\footnotesize
\caption{\textbf{Golden-context effects and matched controls.} Gold log-probability changes for the same interventions. $R_{\rm only}=R_{20}\setminus I_{20}$; control heads match the layer and discovery prefix-attention mass of $I^\star$. All entries are point estimates.}
\label{tab:app-mask-controls}
\begin{tabular}{@{}lrrrrr@{}}
\toprule
Model & \shortstack{Golden\\restore $I^\star$} & \shortstack{Golden\\mask $I^\star$} & \shortstack{Golden\\mask $R_{\rm only}$} & \shortstack{Short\\mask control} & \shortstack{Long\\mask control}\\
\midrule
Llama-3.1-8B & $-0.023$ & $-0.024$ & $-0.172$ & $-0.020$ & $-0.046$\\
Mistral-7B-v0.1 & $-0.003$ & $-0.008$ & $-0.089$ & $+0.001$ & $-0.005$\\
Qwen2.5-7B & $-0.019$ & $-0.021$ & $-0.155$ & $+0.001$ & $-0.002$\\
Pythia-6.9B & $-0.041$ & $-0.007$ & $-0.075$ & $+0.015$ & $+0.025$\\
GPT-2 XL & $-0.012$ & $-0.012$ & $-0.189$ & $+0.005$ & $+0.033$\\
\bottomrule
\end{tabular}
\end{table}

\begin{figure}[!htbp]\centering\includegraphics[width=\linewidth]{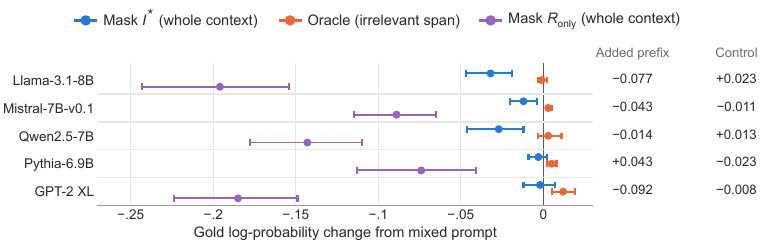}
\caption{\textbf{Useful evidence with distraction.} The irrelevant prefix precedes the golden prefix and query. Points and 95\% intervals compare edits with the unedited mixed prompt; $I^\star$ and $R_{\rm only}$ masks block all context, while the oracle uses the irrelevant-span boundary. The added-prefix column compares mixed with golden alone. The matched-control column is a point estimate relative to mixed.}\label{tab:app-mixed-evidence}\label{fig:app-mixed-evidence}
\end{figure}

When irrelevant and golden prefixes occur together, whole-context masking and oracle masking target different spans (Figure~\ref{tab:app-mixed-evidence}). The whole-context $I^\star$ mask reduces gold log-probability in Llama, Mistral, and Qwen, while the oracle restricts the edit to the known irrelevant span.

\subsection{Direct prefix access and the copier-to-query test}
\label{sec:app-intrusion-route}
\begin{figure}[!htbp]\centering\includegraphics[width=\linewidth]{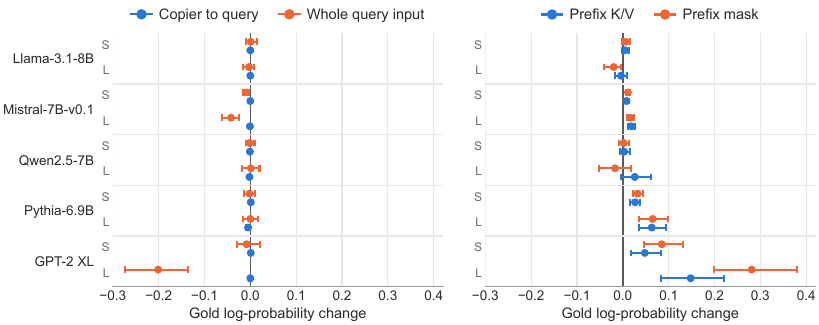}
\caption{\textbf{Query-input and prefix-access interventions.} Gold log-probability changes with 95\% intervals. Query substitutions use bare-query donors; prefix K/V uses token-length-matched filler. END is unpatched in the K/V intervention. Both panels use the same axis. Small changes in total prefix attention do not imply fixed source-wise attention.}\label{tab:app-intrusion-route}\label{fig:app-intrusion-route}
\end{figure}

The tested copier-to-query substitution changes gold log-probability by $-0.005$ to $+0.001$. Whole-query donors and aligned filler-prefix K/V replacements provide broader comparisons (Figure~\ref{tab:app-intrusion-route}). K/V substitution recomputes attention and improves gold log-probability in both prefix conditions for Mistral, Pythia, and GPT-2 XL. The experiment isolates the specified copier contribution; the relative query-norm changes are $0.008$--$0.035$ (Table~\ref{tab:app-intrusion-route-attention}).

\begin{table}[!htbp]
\setlength{\tabcolsep}{4pt}
\centering\footnotesize
\setlength{\tabcolsep}{9pt}
\caption{\textbf{Query and attention changes.} Relative query-norm changes and mean END-to-prefix attention-mass changes under the same substitutions. The copier contribution is small relative to the full query, and prefix K/V substitution changes what the heads read with little change in total prefix attention. Values are point estimates.}
\label{tab:app-intrusion-route-attention}
\begin{tabular}{@{}llrrr@{}}
\toprule
 & & \multicolumn{2}{c}{Copier$\to$query} & Prefix K/V\\
\cmidrule(lr){3-4}
Model & Prefix & $\|\Delta q\|/\|q\|$ & $\Delta a_{\rm prefix}$ & $\Delta a_{\rm prefix}$\\
\midrule
Llama-3.1-8B & Short & $0.008$ & $+0.000$ & $+0.002$\\
 & Long & $0.010$ & $+0.001$ & $-0.001$\\
Mistral-7B-v0.1 & Short & $0.012$ & $+0.001$ & $+0.001$\\
 & Long & $0.013$ & $+0.002$ & $-0.001$\\
Qwen2.5-7B & Short & $0.018$ & $+0.001$ & $-0.000$\\
 & Long & $0.023$ & $+0.001$ & $-0.002$\\
Pythia-6.9B & Short & $0.025$ & $+0.001$ & $+0.001$\\
 & Long & $0.027$ & $+0.005$ & $+0.002$\\
GPT-2 XL & Short & $0.025$ & $+0.000$ & $+0.000$\\
 & Long & $0.035$ & $+0.001$ & $+0.005$\\
\bottomrule
\end{tabular}
\end{table}

\subsection{Entrainment heads}
\label{sec:app-entrainment}
In Llama-3.1-8B, neither the published 36-head country--capital entrainment set \citep{niu2025llama} nor its 29-head reproduction overlaps the top twenty intrusion heads. The uniform expected overlap is 0.70 (Table~\ref{tab:app-entrainment-overlap}). Their depth profiles differ: 17 of 20 intrusion heads and 3 of 36 published entrainment heads lie in the last third.

\begin{table}[!htbp]
\setlength{\tabcolsep}{4pt}
\centering\footnotesize
\caption{\textbf{Overlap with the published entrainment set.} Shared heads at each cutoff of the Llama-3.1-8B intrusion ranking; expected counts use uniformly sampled head sets.}
\label{tab:app-entrainment-overlap}
\begin{tabular}{@{}lrrrrrrr@{}}
\toprule
Top intrusion $k$ & 5 & 10 & 20 & 40 & 60 & 100 & 150\\
\midrule
Shared heads & 0 & 0 & 0 & 1 & 3 & 5 & 7\\
Expected & 0.18 & 0.35 & 0.70 & 1.41 & 2.11 & 3.52 & 5.27\\
\bottomrule
\end{tabular}
\end{table}

\subsection{Contextual retrieval}
\label{sec:app-contextual-retrieval}
Retrieval ranking uses attention to the golden answer position weighted by the gold-directed write. The global top-twenty rankings overlap in 3, 5, 6, 5, and 3 heads, respectively. Table~\ref{tab:app-retrieval-mask} compares the resulting $I_{\rm only}$ and $R_{\rm only}$ sets, not the twenty-head $I^\star$ population. Their sizes are 14--17 after exclusion. Intrusion-only masking increases the gold logit in Llama, Mistral, Pythia, and GPT-2 XL under both irrelevant conditions; Qwen's changes are negative. Golden-context changes are reported alongside the irrelevant-context changes in the same logit units.

\begin{table}[!htbp]
\setlength{\tabcolsep}{4pt}
\centering\footnotesize
\caption{\textbf{Contextual retrieval and intrusion.} Gold-logit change under END-to-prefix masking, relative to the unedited condition. Both exclusive populations contain $n$ heads per model; random sets match this size and the intrusion-only layer distribution.}
\label{tab:app-retrieval-mask}
\begin{tabular}{@{}llrrr@{}}
\toprule
Model & Masked population & Golden & Short irrelevant & Long irrelevant\\
\midrule
Llama-3.1-8B & Intrusion-only & $-0.056$ & $+0.052$ & $+0.070$\\
($n=17$) & Retrieval-only & $-0.494$ & $+0.027$ & $+0.028$\\
 & Random & $+0.042$ & $-0.002$ & $-0.002$\\[3pt]
Mistral-7B-v0.1 & Intrusion-only & $-0.083$ & $+0.001$ & $+0.054$\\
($n=15$) & Retrieval-only & $-0.276$ & $+0.022$ & $+0.031$\\
 & Random & $-0.015$ & $-0.009$ & $-0.029$\\[3pt]
Qwen2.5-7B & Intrusion-only & $-0.233$ & $-0.107$ & $-0.030$\\
($n=14$) & Retrieval-only & $-0.370$ & $-0.004$ & $+0.010$\\
 & Random & $+0.016$ & $-0.007$ & $+0.001$\\[3pt]
Pythia-6.9B & Intrusion-only & $-0.108$ & $+0.159$ & $+0.087$\\
($n=15$) & Retrieval-only & $-0.068$ & $+0.068$ & $+0.035$\\
 & Random & $+0.026$ & $+0.050$ & $+0.004$\\[3pt]
GPT-2 XL & Intrusion-only & $-0.034$ & $+0.150$ & $+0.369$\\
($n=17$) & Retrieval-only & $-0.299$ & $+0.022$ & $+0.087$\\
 & Random & $-0.002$ & $-0.018$ & $+0.010$\\
\bottomrule
\end{tabular}
\end{table}
\section{Downstream Evaluation}
\label{sec:appendix-downstream}\label{sec:appendix}

We compare clean accuracy $C$, distracted accuracy $D$, and edited accuracy $E$ for oracle residual patching and the K10/K20 head masks. K10 denotes ten intrusion-ranked heads after retrieval-head exclusion; K20 denotes the full top-20 ranking, including overlapping heads. These are separate from the 20-head disjoint population and the exclusive-set comparisons in Appendix~C.

Oracle patching replaces the END-position \texttt{resid\_pre} at each decoding step with a matched clean-query state. The reported layers are $L=30$ for Llama-3.1-8B, Mistral-7B-v0.1 and Pythia-6.9B, and $L=26$ for Qwen2.5-7B. The larger models use layer-swept settings. This intervention provides a clean-state recovery reference.

\begin{table}[!htbp]
\centering
\footnotesize
\setlength{\tabcolsep}{6pt}
\caption{Cross-task overlap of intrusion-head sets. The reference is the same model's factual-recall ten-head set; each entry is $\lvert H^{(10)}_{\mathrm{factual}}\cap H^{(10)}_{\mathrm{task}}\rvert/10$. $^{*}$Reference set.}
\label{tab:cross-task}
\begin{tabular}{@{}lrrrr@{}}
\toprule
\textbf{Task} & \textbf{Llama} & \textbf{Mistral} & \textbf{Qwen} & \textbf{Pythia} \\
\midrule
Factual recall$^{*}$ & 10/10 & 10/10 & 10/10 & 10/10 \\
IOI (retrieval) & 10/10 & 10/10 & 9/10 & 10/10 \\
GSM-DC & 10/10 & 9/10 & 9/10 & 10/10 \\
GSM-IC & 10/10 & 9/10 & 9/10 & 10/10 \\
HotpotQA & 10/10 & 10/10 & 10/10 & 10/10 \\
PubMedQA & 10/10 & 10/10 & 10/10 & 10/10 \\
\bottomrule
\end{tabular}
\end{table}

\subsection{GSM-DC}
\label{sec:appendix-gsmdc}
\begin{figure}[!htbp]\centering
\includegraphics[width=\linewidth]{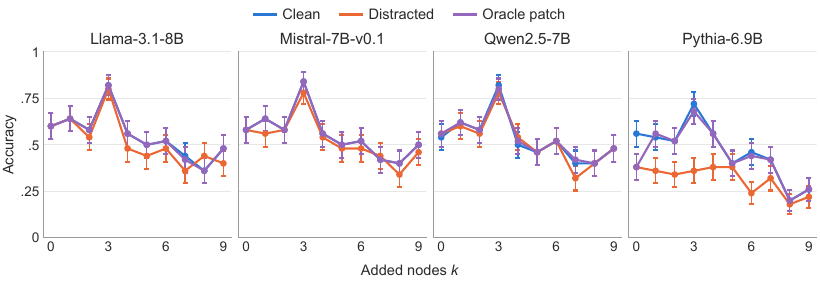}
\caption{GSM-DC accuracy at the ten plotted settings $k=0,\ldots,9$.}
\label{fig:gsmdc-permodel}\end{figure}

Figure~\ref{fig:gsmdc-permodel} reports GSM-DC accuracy by model \citep{yang2025gsmdc}. The clean baseline varies across $k$; comparisons are therefore made within each plotted condition. The figure supplies clean, distracted and patch outcomes.

\subsection{GSM-IC: Single-Token Evaluation}
\label{sec:appendix-gsmic-singletoken}
The single-token evaluation measures argmax correctness at the first answer token on GSM-IC \citep{shi2023distracted}. Patched accuracy equals or exceeds clean accuracy in each row (Table~\ref{tab:gsmic-singletoken}).

\begin{table}[!htbp]\centering\small\setlength{\tabcolsep}{6pt}
\caption{Single-token accuracy. Gain is $100(E-D)$ percentage points (pp). Recovery uses the clean--distracted gap; a value above 100\% corresponds to $E>C$.}\label{tab:gsmic-singletoken}
\begin{tabular}{@{}lrrrrr@{}}\toprule
Model & Clean & Distracted & Patch & Patch gain (pp) & Recovery (\%) \\\midrule
Llama-3.1-70B & .578 & .408 & .656 & +24.8 & 145.9 \\
Qwen2.5-72B & .900 & .840 & .932 & +9.2 & 153.3 \\
Qwen2.5-32B & .888 & .810 & .924 & +11.4 & 146.2 \\
Qwen2.5-7B & .688 & .638 & .730 & +9.2 & 184.0 \\
Mistral-7B-v0.1 & .456 & .352 & .456 & +10.4 & 100.0 \\
\bottomrule\end{tabular}
\end{table}

\subsection{GSM-IC: Chain-of-Thought Evaluation}
\label{sec:appendix-gsmic-cot}
K10 improves accuracy by 13 and 12 percentage points for Mistral and Llama, respectively; K20 improves it by 4 and 7 points (Table~\ref{tab:gsmic-cot}).

\begin{table}[!htbp]\centering\small\setlength{\tabcolsep}{7pt}
\caption{GSM-IC chain-of-thought accuracy. Relative to the distractor-induced loss, K10/K20 recover 46.4\%/14.3\% on Mistral and 70.6\%/41.2\% on Llama. These percentages use $C-D$, not the gain from oracle patching.}\label{tab:gsmic-cot}
\begin{tabular}{@{}lrrrrr@{}}\toprule
Model & Clean & Distracted & Patch & K10 & K20 \\\midrule
Mistral-7B-v0.1 & .66 & .38 & .66 & .51 & .42 \\
Llama-3.1-8B & .64 & .47 & .67 & .59 & .54 \\
\bottomrule\end{tabular}
\end{table}

\paragraph{Accuracy changes and recovery.}
For every observed condition we report $\Delta\mathrm{acc}=100(E-D)$ in percentage points. Conditional loss recovery is
\begin{equation}
R=100\frac{E-D}{C-D},\qquad C>D.
\label{eq:downstream-recovery}
\end{equation}
When $C\leq D$, the condition remains in the absolute-effect analysis but has no loss-recovery value. Table~\ref{tab:downstream-recomputed} compares all three interventions on the same observed cells within each task. Each cell receives equal weight.

\begin{table}[!htbp]\centering\footnotesize\setlength{\tabcolsep}{5pt}
\caption{Common-cell summaries: $n$ counts cells with all three intervention results; $n_+$ further requires $C>D$.}\label{tab:downstream-recomputed}
\begin{tabular}{@{}lcrrrcrrr@{}}\toprule
 & & \multicolumn{3}{c}{Mean change (pp)} & & \multicolumn{3}{c}{Mean loss recovery (\%)} \\
Task & $n$ & Patch & K10 & K20 & $n_+$ & Patch & K10 & K20 \\\midrule
GSM-IC CoT & 2 & +24.00 & +12.50 & +5.50 & 2 & 108.82 & 58.51 & 27.73 \\
HotpotQA & 15 & +23.07 & +14.53 & +9.27 & 13 & 85.51 & 48.54 & 29.53 \\
PubMedQA & 11 & +15.64 & +5.45 & +3.09 & 9 & 99.26 & 45.31 & 17.67 \\
\bottomrule\end{tabular}
\end{table}

\subsection{HotpotQA}
\label{sec:appendix-hotpotqa}
\begin{figure}[!htbp]\centering
\includegraphics[width=\linewidth]{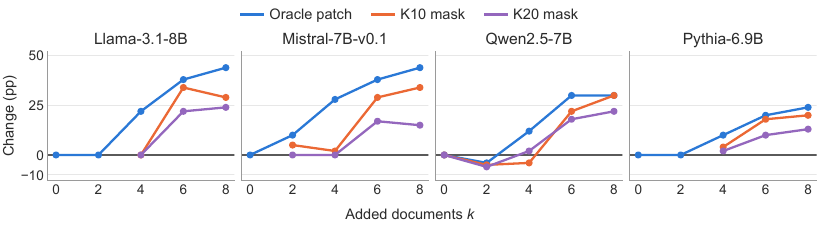}
\caption{Edited-minus-distracted exact-match accuracy on HotpotQA. Lines connect observed conditions. All panels use the same scale, and zero denotes no change.}\label{fig:def-hotpot-effects}
\end{figure}

HotpotQA \citep{yang2018hotpotqa} uses two supporting passages and $k$ added non-supporting paragraphs. At $k=8$, K10 gains 29, 34, 30 and 20 percentage points on Llama, Mistral, Qwen and Pythia, respectively. Figure~\ref{fig:def-hotpot-effects} shows the full response across $k$, including unchanged and adverse effects.
\begin{table}[!htbp]\centering\footnotesize\setlength{\tabcolsep}{6pt}\renewcommand{\arraystretch}{1.03}
\caption{HotpotQA exact-match accuracy.}\label{tab:hotpotqa}
\begin{tabular}{@{}llrrrrr@{}}\toprule
Model & Condition & $k=0$ & $k=2$ & $k=4$ & $k=6$ & $k=8$ \\\midrule
Llama-3.1-8B & Clean & 0.68 & 0.68 & 0.68 & 0.68 & 0.68 \\
 & Distracted & 0.68 & 0.70 & 0.44 & 0.22 & 0.18 \\
 & Oracle patch & 0.68 & 0.70 & 0.66 & 0.60 & 0.62 \\
 & K10 mask & --- & --- & 0.44 & 0.56 & 0.47 \\
 & K20 mask & --- & --- & 0.44 & 0.44 & 0.42 \\
\midrule
Mistral-7B & Clean & 0.72 & 0.72 & 0.72 & 0.72 & 0.72 \\
 & Distracted & 0.72 & 0.62 & 0.40 & 0.30 & 0.24 \\
 & Oracle patch & 0.72 & 0.72 & 0.68 & 0.68 & 0.68 \\
 & K10 mask & --- & 0.67 & 0.42 & 0.59 & 0.58 \\
 & K20 mask & --- & 0.62 & 0.40 & 0.47 & 0.39 \\
\midrule
Qwen2.5-7B & Clean & 0.72 & 0.72 & 0.72 & 0.72 & 0.72 \\
 & Distracted & 0.72 & 0.74 & 0.50 & 0.26 & 0.22 \\
 & Oracle patch & 0.72 & 0.70 & 0.62 & 0.56 & 0.52 \\
 & K10 mask & 0.72 & 0.69 & 0.46 & 0.48 & 0.52 \\
 & K20 mask & 0.72 & 0.68 & 0.52 & 0.44 & 0.44 \\
\midrule
Pythia-6.9B & Clean & 0.42 & 0.42 & 0.42 & 0.42 & 0.42 \\
 & Distracted & 0.42 & 0.42 & 0.32 & 0.22 & 0.18 \\
 & Oracle patch & 0.42 & 0.42 & 0.42 & 0.42 & 0.42 \\
 & K10 mask & --- & --- & 0.36 & 0.40 & 0.38 \\
 & K20 mask & --- & --- & 0.34 & 0.32 & 0.31 \\
\bottomrule\end{tabular}
\end{table}

\subsection{PubMedQA}
\label{sec:appendix-pubmedqa}
PubMedQA \citep{jin2019pubmedqa} appends $k$ irrelevant abstracts to the supporting context. At $k=4$, K10 increases accuracy by 7, 21 and 26 percentage points on Llama, Mistral and Qwen. The same comparison includes a $-9$ point effect on Pythia at $k=1$, where the added abstract had improved baseline accuracy.
\begin{figure}[!htbp]\centering
\includegraphics[width=\linewidth]{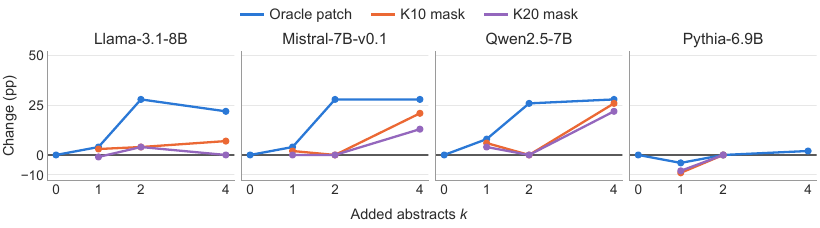}
\caption{Edited-minus-distracted accuracy on PubMedQA, with the same scale as the HotpotQA panels. At Pythia $k=1$, $C=.62<D=.66$; K10/K20 produce .57/.58, corresponding to $-9/-8$ pp, not positive loss recovery.}\label{fig:def-pubmed-effects}
\end{figure}
\begin{table}[!htbp]\centering\footnotesize\setlength{\tabcolsep}{6pt}\renewcommand{\arraystretch}{1.03}
\caption{PubMedQA accuracy. At $k=2$, K10 differs from the distracted baseline on Llama (.54 to .58) and is unchanged on the other three models.}\label{tab:pubmedqa}
\begin{tabular}{@{}llrrrr@{}}\toprule
Model & Condition & $k=0$ & $k=1$ & $k=2$ & $k=4$ \\\midrule
Llama-3.1-8B & Clean & 0.82 & 0.82 & 0.82 & 0.82 \\
 & Distracted & 0.82 & 0.78 & 0.54 & 0.60 \\
 & Oracle patch & 0.82 & 0.82 & 0.82 & 0.82 \\
 & K10 mask & --- & 0.81 & 0.58 & 0.67 \\
 & K20 mask & --- & 0.77 & 0.58 & 0.60 \\
\midrule
Mistral-7B & Clean & 0.86 & 0.86 & 0.86 & 0.86 \\
 & Distracted & 0.86 & 0.82 & 0.58 & 0.58 \\
 & Oracle patch & 0.86 & 0.86 & 0.86 & 0.86 \\
 & K10 mask & --- & 0.84 & 0.58 & 0.79 \\
 & K20 mask & --- & 0.82 & 0.58 & 0.71 \\
\midrule
Qwen2.5-7B & Clean & 0.74 & 0.74 & 0.74 & 0.74 \\
 & Distracted & 0.74 & 0.66 & 0.48 & 0.44 \\
 & Oracle patch & 0.74 & 0.74 & 0.74 & 0.72 \\
 & K10 mask & --- & 0.72 & 0.48 & 0.70 \\
 & K20 mask & --- & 0.70 & 0.48 & 0.66 \\
\midrule
Pythia-6.9B & Clean & 0.62 & 0.62 & 0.62 & 0.62 \\
 & Distracted & 0.62 & 0.66 & 0.62 & 0.60 \\
 & Oracle patch & 0.62 & 0.62 & 0.62 & 0.62 \\
 & K10 mask & --- & 0.57 & 0.62 & --- \\
 & K20 mask & --- & 0.58 & 0.62 & --- \\
\bottomrule\end{tabular}
\end{table}

\section{Base and Instruction-Tuned Checkpoint Comparison}
\label{app:instruct}
We compare Llama-3.1-8B and Llama-3.1-8B-Instruct on identical raw-continuation prompts, without a chat template. The shared architecture permits comparison of the same layer/head indices. Table~\ref{tab:instruct_population} reports late-block ranking correlations and set overlaps for this checkpoint comparison; it is separate from the global head-set analysis in Appendix~C.

\begin{table}[!htbp]\centering\small\setlength{\tabcolsep}{8pt}
\caption{Checkpoint-level population comparison. Correlations compare per-head scores; Jaccard values compare selected memberships. These statistics do not by themselves establish transfer of a frozen base-model mask.}\label{tab:instruct_population}
\begin{tabular}{@{}lrr@{}}\toprule
Metric & Base & Instruct / comparison \\\midrule
$|R\cap I|$, top-20 late-block sets & 5/20 & 5/20 \\
Spearman, intrusion-score rankings & --- & .924 \\
Spearman, retrieval-score rankings & --- & .962 \\
Jaccard, top-20 retrieval sets & --- & .818 \\
Jaccard, top-20 intrusion sets & --- & .481 \\
Jaccard, disjoint top-10 sets & --- & .429 \\
\bottomrule\end{tabular}
\end{table}

\paragraph{Head-mask effects.}
Table~\ref{tab:instruct_knockout} reports the class-conditional interventions for the two checkpoints. The short-prefix K10 changes are +.009 and +.006 logits; the long-prefix changes are +.015 and $-.017$. K20 yields positive irrelevant-context changes in both checkpoints. Bare-query invariance follows from the absence of context edges.

\begin{table}[!htbp]\centering\small\setlength{\tabcolsep}{6pt}
\caption{Gold-logit changes, knockout minus unedited baseline. The golden column uses this same sign convention.}\label{tab:instruct_knockout}
\begin{tabular}{@{}llrrrr@{}}\toprule
Head set & Checkpoint & Short $\Delta\ell_g$ & Long $\Delta\ell_g$ & Golden $\Delta\ell_g$ & Bare \\\midrule
Disjoint K10 & Base & +.009 & +.015 & +.010 & .000 \\
 & Instruct & +.006 & $-.017$ & +.039 & .000 \\
Full K20 & Base & +.149 & +.151 & +.161 & .000 \\
 & Instruct & +.035 & +.048 & +.113 & .000 \\
\bottomrule\end{tabular}
\end{table}

\paragraph{Oracle patching across tasks.}
With residual patching at $L=30$, the instruction-tuned checkpoint recovers 94.7--100.0\% of the clean--distracted gap on the four task summaries (Table~\ref{tab:instruct_downstream}). This comparison evaluates oracle state substitution.

\begin{table}[!htbp]\centering\small\setlength{\tabcolsep}{9pt}
\caption{Task summaries for oracle patching at $L=30$, averaged over conditions with $C>D$. The base GSM-IC denominator is approximately zero.}\label{tab:instruct_downstream}
\begin{tabular}{@{}lrr@{}}\toprule
Task & Base recovery (\%) & Instruct recovery (\%) \\\midrule
GSM-IC, single-token & Undefined & 94.7 \\
GSM-DC & 87.5 & 100.0 \\
HotpotQA & 89.8 & 97.7 \\
PubMedQA & 100.0 & 100.0 \\
\bottomrule\end{tabular}
\end{table}

\section{Sensitivity to Head Count and Retrieval Penalty}
\label{app:ksweep}
We vary the head count $K\in\{5,8,10,12,15,20\}$ and introduce a retrieval penalty $\alpha_{\mathrm{sel}}\in\{0,.5,1,2\}$ in the ranking
\begin{equation}
s_{\mathrm{sel}}(h)=z\!\left(s_{\mathrm{int}}(h)\right)-\alpha_{\mathrm{sel}}\,z\!\left(s_{\mathrm{ret}}(h)\right).
\label{eq:def-selection-penalty}
\end{equation}
Here $z$ standardizes each score across candidate heads. The baseline ranking uses $\alpha_{\mathrm{sel}}=0$; this penalty is separate from any restoration-strength parameter. We plot short-irrelevant gold-logit changes and the accompanying golden-context scores as separate quantities.

\subsection{Head-Count Sweep}
\begin{figure}[!htbp]\centering
\includegraphics[width=\linewidth]{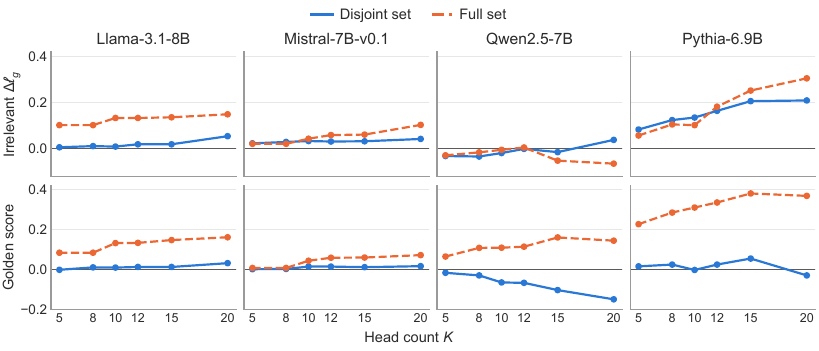}
\caption{Head-count sweep at zero retrieval penalty. Within each model, the upper panel shows short-irrelevant $\Delta\ell_g$ (edited minus baseline); the lower panel shows the golden-context score.}\label{tab:ksweep}
\end{figure}

The disjoint and full rankings yield different response curves (Figure~\ref{tab:ksweep}). The disjoint-set short-prefix change at $K=20$ is +.054, +.042, +.038 and +.209 logits for Llama, Mistral, Qwen and Pythia. The complete curves retain the changes of direction between adjacent values of $K$.

\subsection{Joint Head-Count and Penalty Sweep}
\begin{figure}[!htbp]\centering
\includegraphics[width=\linewidth]{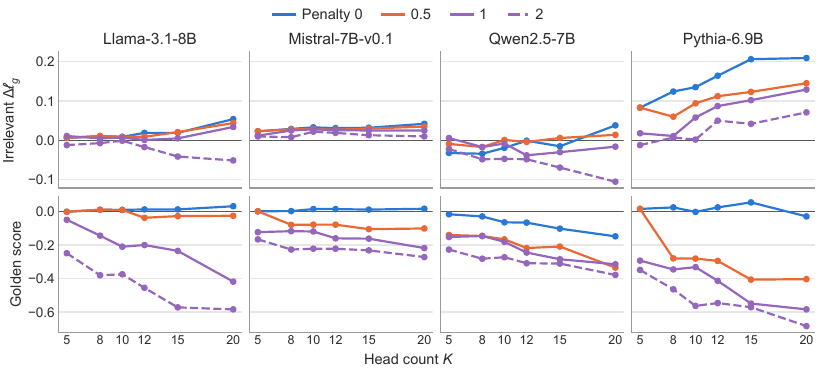}
\caption{Complete $K\times\alpha_{\mathrm{sel}}$ grid for the disjoint ranking. Blue, orange and purple distinguish penalties; the dashed purple curve denotes penalty 2. Upper panels show short-prefix $\Delta\ell_g$; lower panels show the golden-context score. These are point estimates.}
\label{tab:alphasweep}\label{tab:kalpha_llama}\label{tab:kalpha_mistral}\label{tab:kalpha_qwen}\label{tab:kalpha_pythia}
\end{figure}

Figure~\ref{tab:kalpha_llama} presents every setting. Zero penalty gives the largest $K=10$ short-prefix change on Llama, Mistral and Pythia; Qwen's largest value occurs at penalty .5. The curves describe sensitivity over the evaluated settings, with no smoothing or removal of negative outcomes.

\end{document}